\documentclass{article} 
\usepackage{iclr2027_conference,times}

\iclrfinalcopy

\usepackage[utf8]{inputenc}
\usepackage[T1]{fontenc}
\usepackage{hyperref}
\usepackage{url}
\usepackage{booktabs}
\usepackage{amsfonts}
\usepackage{amsmath}
\usepackage{amssymb}
\usepackage{graphicx}
\usepackage{algorithm}
\usepackage{multirow}
\usepackage{algorithmic}
\usepackage{float}
\usepackage{tikz}
\usetikzlibrary{arrows.meta, calc, positioning, backgrounds,
  fit, shapes.geometric, fadings, decorations.markings}
\usepackage{subcaption}
\usepackage{siunitx}

\definecolor{attnFill}{RGB}{218,231,246}
\definecolor{attnHi}{RGB}{234,241,251}
\definecolor{attnStroke}{RGB}{90,133,192}
\definecolor{attnText}{RGB}{46,74,120}
\definecolor{mlpFill}{RGB}{250,240,216}
\definecolor{mlpHi}{RGB}{254,248,236}
\definecolor{mlpStroke}{RGB}{200,160,48}
\definecolor{mlpText}{RGB}{122,100,32}
\definecolor{embFill}{RGB}{234,234,234}
\definecolor{embHi}{RGB}{245,245,245}
\definecolor{embStroke}{RGB}{170,170,170}
\definecolor{embText}{RGB}{96,96,96}
\definecolor{ghostFill}{RGB}{230,236,245}
\definecolor{ghostStroke}{RGB}{138,164,204}
\definecolor{ghostText}{RGB}{136,152,184}
\definecolor{tgtFill}{RGB}{238,232,248}
\definecolor{tgtHi}{RGB}{248,244,255}
\definecolor{tgtStroke}{RGB}{136,120,168}
\definecolor{tgtText}{RGB}{88,72,160}
\definecolor{shCol}{RGB}{200,200,204}
\definecolor{ruleG}{RGB}{232,232,232}
\definecolor{labG}{RGB}{187,187,187}
\definecolor{layG}{RGB}{204,204,204}
\definecolor{tokC}{RGB}{72,72,72}
\definecolor{posC}{RGB}{176,176,176}

\tikzset{
  cb/.style={minimum width=1.8cm, minimum height=0.62cm,
    rounded corners=3pt, inner sep=0pt, line width=0.5pt,
    font=\sffamily\scriptsize\bfseries, align=center},
  cs/.style={minimum width=1.8cm, minimum height=0.62cm,
    rounded corners=3pt, fill=shCol, opacity=0.22},
  atn/.style={cb, top color=attnHi, bottom color=attnFill,
    draw=attnStroke, text=attnText},
  mlp/.style={cb, top color=mlpHi, bottom color=mlpFill,
    draw=mlpStroke, text=mlpText},
  emb/.style={cb, top color=embHi, bottom color=embFill,
    draw=embStroke, text=embText},
  gho/.style={cb, fill=ghostFill, draw=ghostStroke,
    text=ghostText, dash pattern=on 2.5pt off 1.8pt,
    draw opacity=0.55, text opacity=0.7},
  tgt/.style={cb, top color=tgtHi, bottom color=tgtFill,
    draw=tgtStroke, text=tgtText, line width=0.7pt},
  ll/.style={font=\sffamily\scriptsize, text=layG, anchor=east},
  tl/.style={font=\sffamily\small\bfseries, text=tokC, anchor=north},
  pl/.style={font=\sffamily\tiny, text=posC, anchor=north},
  mt/.style={font=\sffamily\tiny\bfseries, opacity=0.8},
  ar/.style={-{Stealth[length=3.5pt, width=2.8pt]}},
  dar/.style={ar, dash pattern=on 4pt off 2.5pt},
}

\title{Every Component Is a Lookup:\\ One Linear Graph for Interaction, Composition and Attribution}

\author{Po-Kai Chen \\
Leiden University \\
\texttt{pokaichen.ai@gmail.com}
\And
Aske Plaat \\
Leiden University \\
\texttt{a.plaat@liacs.leidenuniv.nl}
\And
Niki van Stein \\
Leiden University \\
\texttt{n.van.stein@liacs.leidenuniv.nl}
}

\begin{document}

\maketitle
\lhead{Preprint} 

\begin{abstract}

Interpretability methods for transformers are typically built around separate questions: which components interact, how information routes to the output, and which input tokens contribute. Because these methods rely on different assumptions, their answers are difficult to relate. We argue that two architecturally motivated assumptions suffice to address all three questions: \textbf{attention and MLPs share a key-value form}, $\phi(S)\,U$, in which $\phi(S)$ selects over values $U$, and \textbf{components read from an additive residual stream}, the sum of component outputs. Holding these selections at their forward-pass values turns the model into a computational graph, of which component interactions, composition paths, and token attribution are different readouts. We develop \textsc{Unpack}, a backward attribution procedure over this graph, and validate each readout against the corresponding established test: interaction scores predict ablation effects across models from 160M to 6.9B parameters, recovered routes reproduce established circuits down to the key, query, or value branch the circuit specifies, and token attribution passes the same faithfulness test as dedicated attribution methods. The results suggest that these two assumptions suffice for the interpretability questions above. On a task with a known circuit, we find that contribution and causal effect can differ, and that the difference has a recognisable signature: components that matter for the task change their contribution when the task is removed from the input, while components that act like a bias term do not. Code is available at \url{https://github.com/Fun-Cry/unpacklm}.

\end{abstract}
\section{Introduction}
\label{sec:intro}

In mechanistic interpretability, we are interested in three main questions about a single prediction. (1) Which components feed which, and how strongly? (2) How do those connections compose into the route that produced the answer? (3) Which input tokens drove that answer, and which pushed against it? Each question currently has its own tool, and each tool its own assumptions: path patching establishes which components feed which by intervening with a designed counterfactual prompt~\citep{wang2023ioi}, Information Flow Routes trace the edges between components through a chosen proximity rule~\citep{ferrando-voita-2024-information}, and LRP methods score the input tokens through heuristic propagation rules~\citep{bachlrp, pmlr-v235-achtibat24a}.
In this paper we show how a single linear graph can unify these three views.

The separate assumptions are problematic. Take the LRP family for example: different propagation rules give different attribution scores, so one can neither say what a score means nor which rule is better. And the problem compounds across methods: a method that scores tokens and one that scores components cannot be compared at all; There is no common quantity that both methods measure. Interventional methods avoid this problem because a patched effect has a causal interpretation, but they require a specific counterfactual for each question, and not every prompt supports one.

Motivated by these cases, we argue that two architecturally motivated assumptions suffice to answer all three questions from a single computational graph: both MHA and MLP follow the same key-value template, and both read from a linear residual stream. Self-attention computes $\phi(S)\,U$, a selection $\phi(S)$ over values $U$, with bilinear scores $S = QK^\top/\sqrt{d_\text{head}}$, while an MLP has the same form, with the up-projection's columns as keys, the activation as selection, and the down-projection's rows as values \citep{geva2021transformer}. Because the residual stream is a sum of component outputs \citep{elhage2021mathematical}, holding the routing weights $\phi(S)$ fixed at their forward-pass values makes every score decompose additively over the components that wrote the stream. This yields a direct linear connection between any pair of components, turning the model into a computational graph: connection strengths answer \emph{which components feed which}, paths answer \emph{how those connections compose}, and the credit the paths deposit at the input embeddings answers \emph{which inputs support or oppose the prediction}. The two assumptions determine the structure of this graph completely; the remaining choices of credit normalization and dispatch are stated in Section~\ref{sec:limitations}, and Appendix~\ref{app:beta} shows the results are stable across them.

We develop this view into an algorithm, \textsc{Unpack} (\textbf{Un}ified \textbf{P}ath \textbf{A}ttribution through \textbf{C}omponent \textbf{K}eys), a backward walk over the graph on a single forward pass, with nothing patched, no gradients, and no training, and validate its three outputs against the external checks each purpose-built tool is trusted for: the interaction scores against causal ablation across the Pythia-deduped family (Section~\ref{sec:eval_knockout}), the routes against two circuits established by path patching (Section~\ref{sec:eval_composition} and Appendix~\ref{app:greater_than}), and the token attribution against an input-flipping perturbation check (Section~\ref{sec:eval_tokens}). The graph reads out contribution under the realized computation; Section~\ref{sec:limitations} discusses where contribution and causal effect diverge. Our contributions are:
\begin{itemize}
\item \textbf{A claim and its validation.} We argue that two architecturally motivated assumptions, the shared key-value template and the linear residual stream, are sufficient to answer all three questions from a single computational graph. We validate each readout against its purpose-built counterpart: interaction scores predict causal ablation effects across the Pythia-deduped family (within-layer Spearman $\rho = 0.72$--$0.96$), routes recover two circuits established by path patching down to the key, query, or value branch the circuit names, and token attribution passes the same perturbation check as dedicated attribution methods, using identical hyperparameters throughout.
\item \textbf{An algorithm.} \textsc{Unpack} constructs the graph from a single forward pass: no patching, gradients, or training are required. The graph can be rerooted at any component, providing all three readouts from one run.
\item \textbf{Findings.} Observational credit alone does not give causal effect: layer-0 components receive high credit on every prompt, task or not, and this only becomes visible by comparing against a matched counterfactual, on which credit to the causally established components collapses while layer-0 credit stays put. The credit nonetheless carries mechanism, not just magnitude: on IOI, token attribution tracks a token's circuit role rather than its identity, as the recovered paths predict, and this holds from 160M to 6.9B parameters.
\end{itemize}
\section{Background and Related Work}
\label{sec:related}

\paragraph{Setting and notation.}
A decoder-only Transformer~\citep{attention_origin} stacks $L$
blocks, each with a multi-head self-attention (MHA) sublayer and an MLP
sublayer, both reading from and writing to a residual stream of
width $d_\text{model}$. Head $h$ at query position $q$ forms weights
$\alpha_{h,q,s}$ by softmax over the scaled scores
$\langle X_q, X_s\rangle_{\text{QK}} :=
(X_q W_Q)(X_s W_K)^\top / \sqrt{d_\text{head}}$
and writes $\sum_s \alpha_{h,q,s}\, X_s W_V W_O$; the MLP applies
$W_\text{down}\,\phi(W_\text{up} X_q)$ at each position. We assume
the pre-norm placement used by GPT-2 and Pythia, with LayerNorm on
each sublayer's input.

\paragraph{Residual stream.}
Each sublayer adds its output to the stream, so the residual state
at any layer is a sum, $X_j = \sum_k c_{kj}$, where each $c_{kj}$
is the output of one component at position $j$: an attention head,
an MLP, or the input embedding~\citep{elhage2021mathematical}.
Each component's contribution is a vector in
$\mathbb{R}^{d_\text{model}}$ that can be examined on its own,
which is the property post-hoc interpretability of transformers
generally rests on.

\paragraph{Token attribution.}
Token attribution asks, for a single prediction, how much each input
token contributed to it: one score per input position, signed if the
method distinguishes tokens that pushed the prediction up from
tokens that pushed it down. Gradient- and attention-based methods
include integrated gradients and attention
rollout~\citep{integrated_gradient, abnar-zuidema-2020-quantifying};
relevance-based methods redistribute a conserved score backward
through the network by per-layer
rules~\citep{bachlrp, Chefer_2021_CVPR, ali2022xaitransformersbetterexplanations}.
Section~\ref{sec:eval_tokens} runs our token readout through the
input-flipping check these methods are evaluated on, alongside
three of them: AttnLRP~\citep{pmlr-v235-achtibat24a}, which derives
propagation rules through the softmax and the bilinear attention
product; its variant CP-LRP, which detaches the gradient through
the attention pattern; and Information Flow Routes
(IFR;~\citealp{ferrando-voita-2024-information}), which propagates
$L_1$ mass backward through the residual stream. AttnLRP and
CP-LRP are signed; IFR's per-token scores are non-negative by
construction and do not depend on the target token.
\paragraph{Component attribution and interactions.}
The same question is asked of the model's internal parts: how much
did each head or MLP contribute, and which component feeds which
through which of its inputs. Direct logit
attribution~\citep{nostalgebraist2020logitlens} projects each
component's output onto the target unembedding direction; the
importance our recursion starts from (Eq.~\ref{eq:target_dir}) is
this quantity, which the recursion then passes backward.
Interactions are at present established by intervening: path
patching~\citep{wang2023ioi} overwrites the activation on one edge
with its value on a counterfactual prompt and measures the effect
downstream, and this is how the IOI connections of
Section~\ref{sec:eval_composition}, and the branch each one uses,
were established. Automated and trained variants
exist~\citep{NEURIPS2023_34e1dbe9,
hanna2024faithfaithfulnessgoingcircuit, ameisen2025circuit}.

\section{Method}
\label{sec:method}

\subsection{Two assumptions: linearity and unified key-value framework}
\label{sec:framework}

Our method rests on two structural properties of Transformers: \emph{linearity} in how
sublayer inputs are composed, and a shared \emph{key-value lookup} structure across
MHA and MLPs.

First, the residual stream \citep{elhage2021mathematical} is additive over the outputs of upstream components, $X_j = \sum_k c_{kj}$. The downstream components read it with a linear or bilinear function. For MLP we have
\begin{equation}
    X_q W_{\mathrm{up}} = \sum_k c_{kq} W_{\mathrm{up}}.
\end{equation}
For MHA we have
\begin{equation}
    \langle X_q, X_s\rangle_{\mathrm{QK}}
    = \sum_k \langle X_q,c_{ks}\rangle_{\mathrm{QK}}
    = \sum_k \langle c_{kq},X_s\rangle_{\mathrm{QK}}.
\end{equation}

Second, both MHA and MLPs can be written in a common key-value form (Figure~\ref{fig:unified-kv}) following the interpretation of
\citet{geva2021transformer}. Inside a component, the residual representation $X$ matches each of the keys and produces the score $S_i$, this score is then fed into the nonlinear function $\phi$ to become the weight to select the value $U_i$. Consequently, we can represent both MHA and MLP in one format (Table~\ref{tab:unified}):
\begin{equation}
    \mathrm{Sublayer}(X) = \phi(S)\,U.
\end{equation}

With these assumptions, we can view a transformer model as a computational graph. Between two components, one interacts with the other by contributing to the key-scores $S_i$. Inside a component, each $S_i$ decides how much the corresponding value $U_i$ is selected. To trace the computation backwards, we can reverse the aforesaid order: we first decide the credit of each value $U_i$, the credit is handed to the corresponding score $S_i$, the same credit is then further dispatched among upstream components feeding $S_i$. Section~\ref{sec:recursive} turns this idea into an algorithm and Section~\ref{sec:readouts} shows how this framework and algorithm can answer multiple interpretability questions together.

\begin{table}[h]
\centering
\caption{Unified key-value format $\phi(S)\,U$ for attention and MLP. The entries
$i$ are what the sum $\sum_i \phi(S)_i U_i$ runs over, and are what the recursion of
Section~\ref{sec:recursive} dispatches credit to.}
\label{tab:unified}
\begin{tabular}{lcc}
\toprule
       & Attention & MLP \\
\midrule
$\phi$ & softmax   & ReLU / GELU \\
$S$    & $Q K^T / \sqrt{d_{\text{head}}}$ & $X W_{\text{up}}$ \\
$U$    & $X W_V$   & $W_{\text{down}}$ \\
entries $i$ & source positions $s$ & neurons $j$ \\
\bottomrule
\end{tabular}
\end{table}


\begin{figure}[t]
\centering
\begin{tikzpicture}[
    >=Stealth,
    stream/.style={line width=1.2pt, gray!50},
    stream q/.style={line width=1.6pt, gray!70},
    score box/.style={draw=teal!70!black, fill=teal!#1, minimum width=0.7cm, minimum height=0.55cm, rounded corners=2pt, inner sep=0pt},
    value box/.style={draw=orange!70!black, fill=orange!#1, minimum width=0.55cm, minimum height=0.5cm, rounded corners=2pt, inner sep=0pt},
    karrow/.style={->, teal!60!black, line width=0.6pt},
    varrow/.style={->, orange!70!black, line width=0.6pt},
    qarrow/.style={->, violet!70!black, line width=0.7pt},
    fanline/.style={->, teal!50!black, line width=0.6pt},
    label font/.style={font=\scriptsize\sffamily, text=gray!70!black},
    math font/.style={font=\scriptsize},
]

\begin{scope}[xshift=-3.5cm]

\foreach \i/\lbl in {0/s_1, 1/s_2, 2/s_3} {
    \draw[stream] (\i*1.2, -3.2) -- (\i*1.2, -1.0);
    \node[math font, below] at (\i*1.2, -3.3) {$\lbl$};
    \fill[violet!40] (\i*1.2-0.15, -2.7) rectangle ++(0.3, 0.12);
    \fill[teal!40] (\i*1.2-0.15, -2.55) rectangle ++(0.3, 0.12);
    \fill[orange!40] (\i*1.2-0.15, -2.4) rectangle ++(0.3, 0.12);
}

\draw[stream q] (3.6, -3.2) -- (3.6, -1.0);
\draw[stream q] (3.6, 2.5) -- (3.6, 3.5);
\node[math font, below] at (3.6, -3.3) {$q$};
\fill[violet!40] (3.45, -2.7) rectangle ++(0.3, 0.12);
\fill[teal!40] (3.45, -2.55) rectangle ++(0.3, 0.12);
\fill[orange!40] (3.45, -2.4) rectangle ++(0.3, 0.12);

\foreach \i/\op in {0/10, 1/40, 2/70} {
    \node[score box=\op] (s\i) at (\i*1.2, -0.3) {};
    \node[math font] at (\i*1.2, -0.3) {$\alpha_{\the\numexpr\i+1}$};
}
\node[score box=20] (s3) at (3.6, -0.3) {};
\node[math font] at (3.6, -0.3) {$\alpha_{4}$};
\node[label font, left] at (-0.7, -0.3) {$\varphi(S)\!:$};

\foreach \i in {0,1,2} {
    \draw[karrow] (\i*1.2, -1.0) -- (\i*1.2, -0.7);
}
\draw[karrow] (3.6, -1.0) -- (3.6, -0.7);
\node[label font, text=teal!60!black] at (0.6, -1.35) {$K$};

\draw[violet!50!black, densely dashed, rounded corners=4pt, line width=0.5pt] 
    (-0.5, -0.65) rectangle (4.15, 0.05);

\draw[qarrow, densely dashed] (3.6, -1.6) .. controls (4.6, -1.6) and (4.6, -0.3) .. (4.15, -0.3);
\node[label font, text=violet!60!black] at (4.8, -1.0) {$Q$};

\foreach \i/\op in {0/10, 1/40, 2/70} {
    \node[value box=\op] (v\i) at (\i*1.2, 0.7) {};
    \node[math font] at (\i*1.2, 0.7) {$U_{\the\numexpr\i+1}$};
}
\node[value box=20] (v3) at (3.6, 0.7) {};
\node[math font] at (3.6, 0.7) {$U_{4}$};
\node[label font, left] at (-0.7, 0.7) {$V\!:$};

\draw[varrow, line width={0.3pt}, opacity=0.3] (v0.north) .. controls (0.5, 1.5) .. (3.5, 2.3);
\draw[varrow, line width={0.6pt}, opacity=0.5] (v1.north) .. controls (1.5, 1.6) .. (3.5, 2.2);
\draw[varrow, line width={1.0pt}, opacity=0.8] (v2.north) .. controls (2.5, 1.7) .. (3.5, 2.1);
\draw[varrow, line width={0.4pt}, opacity=0.35] (v3.north) -- (3.6, 2.0);

\node[font=\small\bfseries] at (1.8, 4.0) {Attention};
\end{scope}

\begin{scope}[xshift=3.8cm]

\draw[stream q] (0, -3.2) -- (0, -1.4);
\draw[stream q] (0, 2.5) -- (0, 3.5);
\node[math font, below] at (0, -3.3) {$q$};
\fill[violet!40] (-0.15, -2.7) rectangle ++(0.3, 0.12);
\fill[teal!40] (-0.15, -2.55) rectangle ++(0.3, 0.12);
\fill[orange!40] (-0.15, -2.4) rectangle ++(0.3, 0.12);

\foreach \x in {-1.2, -0.4, 0.4, 1.2} {
    \draw[fanline] (0, -1.4) -- (\x, -0.75);
}

\foreach \x/\op in {-1.2/50, -0.4/15, 0.4/75, 1.2/35} {
    \node[score box=\op] at (\x, -0.3) {};
}
\node[label font, left] at (-1.8, -0.3) {$\varphi(S)\!:$};

\foreach \i/\x/\op in {1/-1.2/50, 2/-0.4/15, 3/0.4/75, 4/1.2/35} {
    \node[value box=\op] (u\i) at (\x, 0.7) {};
    \node[math font] at (\x, 0.7) {$U_{\i}$};
}
\node[label font, left] at (-1.8, 0.7) {$U\!:$};

\draw[varrow, line width={0.7pt}, opacity=0.5] (u1.north) .. controls (-0.6, 1.8) .. (-0.06, 2.35);
\draw[varrow, line width={0.3pt}, opacity=0.2] (u2.north) .. controls (-0.2, 1.7) .. (-0.03, 2.3);
\draw[varrow, line width={1.0pt}, opacity=0.8] (u3.north) .. controls (0.2, 1.7) .. (0.03, 2.3);
\draw[varrow, line width={0.6pt}, opacity=0.4] (u4.north) .. controls (0.6, 1.8) .. (0.06, 2.35);

\node[font=\small\bfseries] at (0, 4.0) {MLP};
\end{scope}

\end{tikzpicture}
\caption{Unified key-value view. Both attention and MLP compute a weighted sum $\sum_i \varphi(S_i) \cdot U_i$: a selection mechanism $\varphi(S)$ (green, opacity indicates strength) weights values $U$ (orange), and the result is added to the residual stream at position~$q$. In attention, position~$q$ serves as both query and key; $K$ and $V$ are indexed by all positions including~$q$, so the output aggregates across positions, while $Q$ reads from~$q$ alone. In MLP, $S$ and $U$ are indexed by neuron, so all computation stays at~$q$. Colored blocks at each position represent upstream residual components~$c_k$; since $S$ decomposes additively across these components, backward credit attribution traces the same paths in reverse.}
\label{fig:unified-kv}
\end{figure}
\subsection{Recursive backward attribution}
\label{sec:recursive}

This algorithm recursively dispatches the credit of the target logit backward, following the mental model of Section~\ref{sec:framework}. Stage 1 and Stage 2 correspond to the key-value and then linearity assumptions. We describe the root and the two stages here; the full algorithm is in Appendix~\ref{app:algorithm}.

\paragraph{Setup.}
At the target position $p$, each residual component $c_k$ has a
scalar contribution to the target logit, obtained by projecting
$c_k$ onto the target unembedding direction:
\begin{equation}
    I(c_k) = \langle c_k(p),\, d_t \rangle,
    \qquad
    d_t := \frac{w_{\text{LN}}}{\sqrt{\text{Var}(X_p) + \epsilon}}
           \odot \bigl(W_U[t] - \overline{W}_U\bigr),
    \label{eq:target_dir}
\end{equation}
where $\overline{W}_U$ is the vocabulary-averaged unembedding
vector.\footnote{The centering makes $I(c_k)$ measure how much
$c_k$ raises the target token's probability rather than its raw
logit, which softmax identifies only up to a constant.} We call
$I(c_k)$ the component's \emph{credit}; it is what the
recursion redistributes.

\paragraph{Stage 1: credit to the entries.}
The entries $i$ indexing $\phi(S)\,U$ are source positions for a
head and neurons for an MLP. Credit $I$
arriving at a component is divided among them by the part of the
output each one supplied along the direction the credit is
travelling in:
\begin{equation}
    I_i \;=\; I \cdot
    \frac{\phi(S)_i\,\langle U_i,\, d \rangle}
         {\textsc{SafeDenom}_k\bigl(\phi(S)_k\,\langle U_k, d\rangle\bigr)},
    \qquad
    d \;=\;
    \begin{cases}
        d_t & \text{at the target (Eq.~\ref{eq:target_dir}),}\\[2pt]
        \phi(S)\,U & \text{deeper in the recursion.}
    \end{cases}
    \label{eq:stage1}
\end{equation}
An entry whose value pushes along $d$ takes positive credit, one
that pushes against it takes negative credit, and an unselected
entry takes none. \textsc{SafeDenom} (App.~\ref{app:safedenom})
floors the denominator so that credit is not amplified without
bound when positive and negative terms cancel.

\paragraph{Stage 2: credit from an entry to the components that built it.}
Two things built entry $i$: the score $S_i$ that selected it, and
the value $U_i$ it holds. Each is additive over the upstream
components that wrote the stream
(Section~\ref{sec:framework}), so $I_i$ is divided among those
components in proportion to their contributions, again through
\textsc{SafeDenom}. The two stages then repeat at each component
so credited, until the credit reaches the embeddings.

For an MLP, the entries are neurons and $U_j = W_{\text{down}}^j$
is a parameter, so only the score carries upstream credit: Stage 2
has a single branch, the writers of the pre-activation at the
MLP's own position.

An attention head reads both $S$ and $U$ from the stream, at two
different positions, so Stage 2 splits three ways, each branch
following one split of the head's inputs:
\begin{itemize}\itemsep1pt \parskip0pt \topsep2pt
\item the \emph{key} branch divides $I_s$ by the key-side split of
  Section~\ref{sec:framework} and continues at $s$;
\item the \emph{query} branch divides $I_s$ by the query-side
  split and continues at $q$;
\item the \emph{value} branch divides $I_s$ by the writers' shares
  of $U_s = X_s W_V W_O$, itself linear in the stream, and
  continues at $s$.
\end{itemize}
These are the Q-, K- and V-composition modes of
\citet{elhage2021mathematical}. We follow all three with equal
weight ($w_K = w_Q = w_V = \tfrac13$) and label each hop with the
branch it took; Appendix~\ref{app:variants} gives the three scores
in full, Figure~\ref{fig:unified-kv} draws the three routes, and
Section~\ref{sec:eval_composition} reads the labels out.

\paragraph{Termination and conservation.}
Source layers strictly decrease at every recursion step, so every
path terminates at the embedding layer, where credit accumulates
at the token position. Aggregate credit is conserved up to the
soft floor on the denominators (Appendix~\ref{app:safedenom}),
which prevents amplification when positive and negative
contributions nearly cancel. We use $\beta = 0.8$ throughout. When enumerating named paths, we drop sub-credits whose magnitude falls below a pruning threshold $\tau = 10^{-3}$; token attribution is computed exactly and does not depend on $\tau$ (Appendix~\ref{app:pseudocode}).

\subsection{Three readouts of the graph}
\label{sec:readouts}

The graph is built once per forward pass. Everything the paper reports is read from it in one of three ways: edge weights between components, paths through the graph, and per-token credit at the input.

\paragraph{Component interaction scores.}
Based on the framework, we can define the interaction score between two components as how strongly one affects the other's selection scores. For MHA the score is a row over source positions $s$. Supposing the downstream head is component $m$ with query position $q$, we define the score as:
\begin{equation}
    \text{score}_{k\to m}^{\text{attn}} = \text{std}(\left\{\langle X_q,c_{ks}\rangle_{\mathrm{QK}}\right\}_s),
    \label{eq:std}
\end{equation}
since softmax is translation-invariant and only the spread of a
row moves the attention pattern.
For an MLP $m$ at position $q$,
the activations are magnitude-sensitive and both signs of a
pre-activation are meaningful, so we use the L2 norm,
\begin{equation}
    \text{score}_{k\to m}^{\text{mlp}} = \|c_{kq} W_{\mathrm{up}}\|_2.
    \label{eq:mlp}
\end{equation}
These are the edge weights of the graph;
Section~\ref{sec:eval_knockout} checks them against ablation.

\paragraph{Named composition paths.}
Recording the path each unit of credit takes from the unembedding back to an embedding gives one path of how the token contributes to the target logit. Each hop names the component and, at attention, the K, Q, or V branch; the credit reaching the embedding scores the
path, e.g.\
$\texttt{emb@S2} \rightarrow \texttt{MLP0@S2} \xrightarrow{V}
\texttt{A8.H6@END}$ (Figure~\ref{fig:composition-paths}). Ranking paths by credit gives the dominant composition routes for a prediction. Since the decomposition is defined at every component, the recursion can be rooted at any of them instead of the target logit (\emph{rerooting}), enumerating the paths into that component.

\paragraph{Token attribution.}
Summing the credit that reaches each input token, and normalizing by
the total positive credit, gives a signed score per token: positive
if the token pushed the target up, negative if it pushed it down.
This is the same quantity as the paths above, summed over all paths
ending at each token.

\section{Experiments and Results}
\label{sec:evaluation}

Do the two assumptions suffice for interpretability tasks? We validate this claim by checking each readout of Section~\ref{sec:readouts} against an established method:
intervention for the edge weights (Section~\ref{sec:eval_knockout}), circuits established by path patching for the paths (Section~\ref{sec:eval_composition}), and input perturbation for the token credit (Section~\ref{sec:eval_tokens}). The decomposition, its
hyperparameters, and the branch weights are identical across all three checks.

\subsection{Component interaction scores predict knockout effects}
\label{sec:eval_knockout}

We test whether the interaction scores of
Section~\ref{sec:readouts} predict what happens when a component's
communication is removed. Over $1{,}000$ Pile
sentences~\citep{pile}, every residual-stream component receives
two aggregated strengths: its summed key-side pull on all
downstream attention heads, and the analogous quantity on all
downstream MLPs. The ablation is communication-specific: the
component's residual vector $c_k$, extracted on a clean forward
pass, is subtracted from the input of one type of downstream
LayerNorm (cut-attention or cut-MLP), and we measure the change in
perplexity ($\Delta\text{ppl}$) on held-out Pile text. Formulas,
sampling, and per-model counts are in
Appendix~\ref{app:knockout_full}.

Figure~\ref{fig:knockout_6_9b} shows the result for
Pythia-6.9B-deduped ($156$ knockouts): the score-to-ablation
relationship is strongly monotonic on both channels, spanning four
to five orders of magnitude on each axis, with within-layer
Spearman $\rho = 0.86$ (attention) and $0.90$ (MLP). The same
pattern holds across the Pythia-deduped
family~\citep{biderman2023pythia} from 160M to 6.9B parameters,
with within-layer $\rho$ between $0.72$ and $0.96$ at every scale
(Appendix~\ref{app:knockout_full}).

\begin{figure}[t]
\centering
\includegraphics[width=0.8\textwidth]{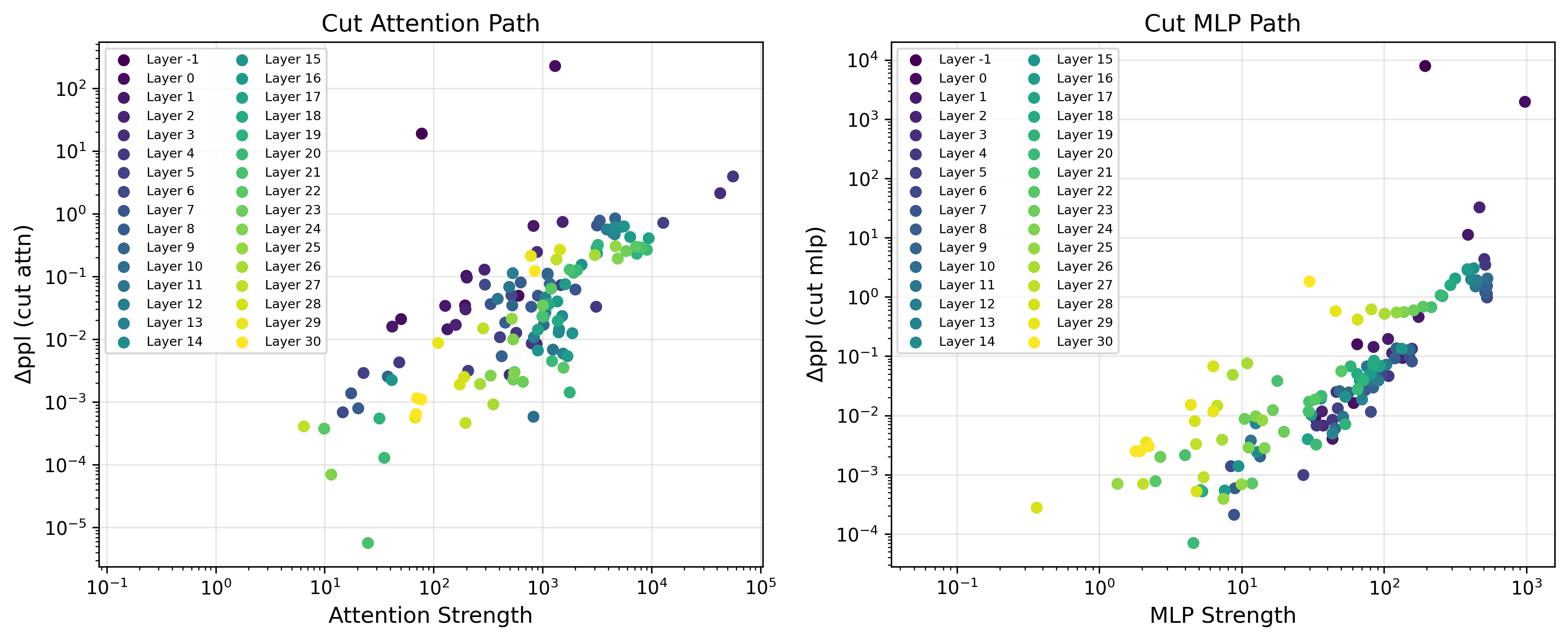}
\caption{Interaction scores against causal knockout effects
(Pythia-6.9B-deduped, 156 knockouts). Each point is one
residual-stream component; the $x$-axis is its aggregated score,
the $y$-axis is $\Delta\text{ppl}$ when the corresponding
communication channel is ablated. \textbf{Left:} attention channel.
\textbf{Right:} MLP channel. Points are colored by source layer
(darker $=$ earlier; layer $-1$ is the embedding); both axes
log-scaled. Within-layer Spearman $\rho = 0.86$ / $0.90$; the same
pattern holds from 160M to 6.9B
(Appendix~\ref{app:knockout_full}).}
\label{fig:knockout_6_9b}
\end{figure}

\subsection{Composition}
\label{sec:eval_composition}

\citet{wang2023ioi} identified a 26-head circuit in GPT-2 small for the IOI task, which asks the model to complete sentences such as ``Alice and Bob went to the store. Bob gave a drink to'' with the name that occurs once (``Alice''). Within the circuit, they name three connections together with the branch each travels: S-Inhibition heads, which suppress the repeated name, modulate the \emph{queries} of Name Mover heads, which copy a name to the output (their Sec.~3.2); Duplicate Token and Induction heads, which detect the repetition, write positional signal into S-Inhibition \emph{values} (Sec.~3.3); and Previous Token heads feed Induction \emph{keys} (Sec.~3.3). We test the compositions found with our method against these claims: over 100 prompts, one decomposition per prompt, we reroot at the head of each claim\footnote{At END for the first two claims, at S2 for the third.} and rank upstream attention heads of layer $\ge 1$ by credit in the claimed branch. We report the median rank of the expected upstream role and found\%, the fraction of rerootings in which it appears at all.

These connections were established by path patching, which measures the causal effect of ablating an edge on a counterfactual prompt; our method is observational, and a component can contribute much while being causally unnecessary, so the two need not agree. However, all three connections surface at the top of the ranking (Table~\ref{tab:comp_ioi}): in the branch each is claimed to use, the named upstream role appears in at least $95\%$ of rerootings with median rank at most $3$. The routing is also branch-specific: at the same Name Mover heads, S-Inhibition is the strongest Q-side input (rank 1, $100\%$) but rank 10 on the K side, the distinction \citet{wang2023ioi} draw and one no single per-token or per-component score can express. Full matrices are in Appendix~\ref{app:composition_matrices}. Appendix~\ref{app:greater_than} repeats the check on the greater-than circuit of \citet{hanna2023greater}.

\begin{table}
\centering
\caption{Composition verification via rerooting on IOI (GPT-2
small, $n=100$). Cells: median rank / found\%, ranking upstream
attention heads (layer $\ge 1$) by credit in the branch of that
row. Bold marks the connection claimed by \citet{wang2023ioi}. The
first two rows reroot at the same Name Mover heads and differ only
in the branch, isolating the mode. Pooled over the heads of each
role ($n=300$ for 3 Name Movers, $n=400$ for 4 S-Inhibition or 4
Induction heads).}
\label{tab:comp_ioi}
\begin{tabular}{llccccc}
\toprule
Branch & Root & NM & S-Inh & Ind & Dup & Prev \\
\midrule
K & NM    & --        & 10 / 48\%          & 14 / 27\%          & 13 / 19\%          & 14 / 46\% \\
Q & NM    & 2 / 33\%  & \textbf{1 / 100\%} & 4 / 63\%           & 18 / 5\%           & 10 / 56\% \\
V & S-Inh & --        & --                 & \textbf{2 / 96\%}  & \textbf{1 / 95\%}  & 10 / 21\% \\
K & Ind   & --        & --                 & 21 / 3\%           & 8 / 10\%           & \textbf{3 / 98\%} \\
\bottomrule
\end{tabular}
\end{table}

\begin{figure}
    \centering
    \includegraphics[width=0.5\linewidth]{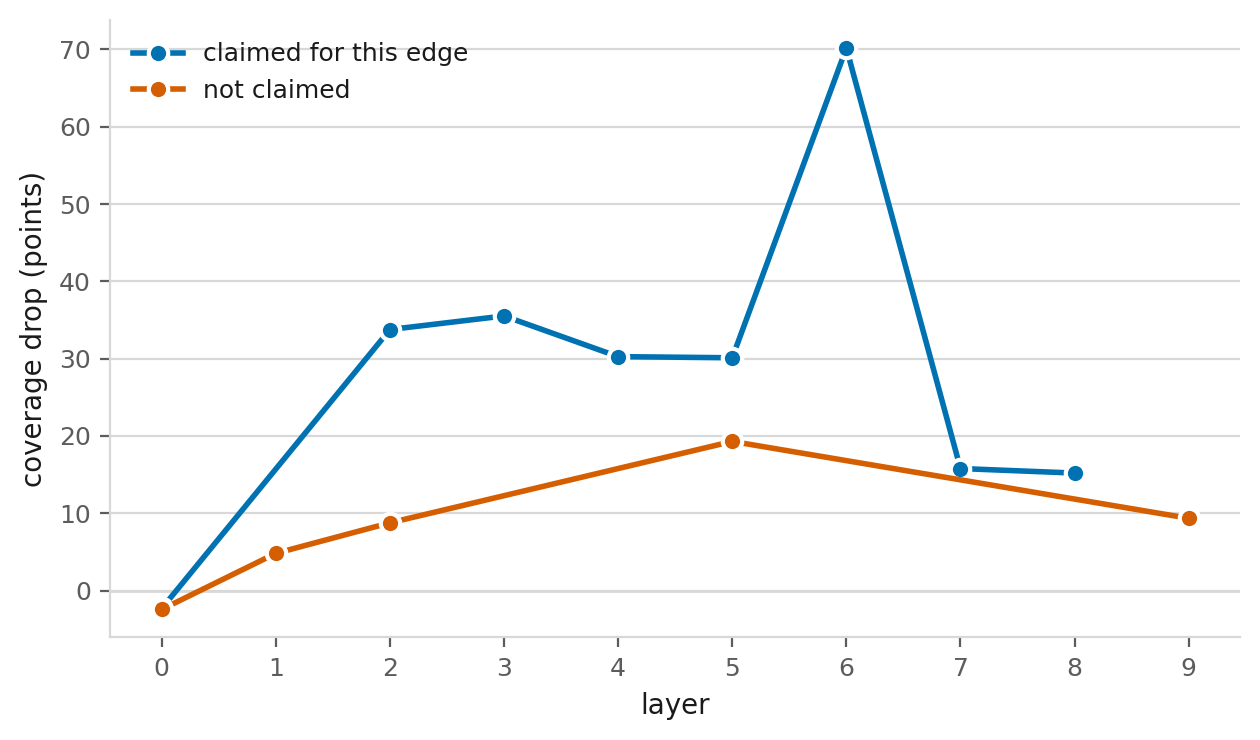}
    \caption{Coverage drop from IOI prompts to matched ABC controls (GPT-2 small), per source layer, pooled over the three claimed edges of Table~\ref{tab:comp_ioi}: how much less often a component appears among the upstream feeders of the claimed branch, re-rootings paired across arms. Controls are the prompts on which the model does not rank A as the top prediction; the claimed line keeps all claimed components, the unclaimed line, the ten strongest unclaimed ones. Claimed components lose $30$--$70$ points at layers 2--6 and about $15$ at layers 7--8, unclaimed at most $19$; at layer 0 both groups barely move, behaving like a bias term. Details in Appendix~\ref{app:coverage}.}
    \label{fig:coverage_drop}
\end{figure}

Some ranks in Table~\ref{tab:comp_ioi} worsen once layer-0
components are included (Appendix~\ref{app:composition_matrices}). We
hypothesize that layer-0 components act like a bias term: they
contribute to every prompt, whether or not the task is running. To check this, we track whether each component still appears among the upstream feeders of the claimed branch when moving from the IOI prompts to the matched ABC controls, and record the \emph{coverage drop}: how much less often it appears when the task is absent (Figure~\ref{fig:coverage_drop}). The components found by path patching drop far more in coverage than the others: their credit is tied to the task. Layer-0 components instead stay put in both prompt sets, keeping their high ranks on any prompt, as a bias would.

\subsection{Token attribution}
\label{sec:eval_tokens}

We evaluate the faithfulness of token attribution with the input flipping protocol of \citet{pmlr-v235-achtibat24a}, reporting Information Flow Routes (IFR), AttnLRP and CP-LRP for comparison. The protocol zeroes input embeddings cumulatively, in most-relevant-first (MoRF) and least-relevant-first (LeRF) order, and scores the area $\Delta A$ between the two target-probability curves (LeRF minus MoRF); \%f is the fraction of sentences with $\Delta A > 0$ ($50\%$ is chance). All four methods clear the check (Table~\ref{tab:faithfulness}; setup and target direction in Appendix~\ref{app:baselines}); \textsc{Unpack} reaches the largest $\Delta A$ ($0.085$) on $100\%$ of sentences, against $-0.001$ ($49\%$) for a random ranking.

\begin{table}
\centering
\caption{Perturbation faithfulness on 300 Pile sentences (GPT-2
small). $\Delta A$ is the area between the LeRF and MoRF curves in
probability units, and \%f the fraction of sentences with
$\Delta A > 0$ ($50\%$ is chance).}
\label{tab:faithfulness}
\begin{tabular}{lcc}
\toprule
Method & $\Delta A$ & \%f \\
\midrule
\textsc{Unpack} & $+0.085$ & 100 \\
AttnLRP         & $+0.066$ &  99 \\
CP-LRP          & $+0.049$ &  91 \\
IFR             & $+0.084$ &  99 \\
\midrule
Random ranking  & $-0.001$ &  49 \\
\bottomrule
\end{tabular}
\end{table}

Intuitively, our token attribution measures the sum of contributions of all paths connecting one input token to the target logit. Since each path records which components compose and through which branch, the token attribution inherits this mechanism. To demonstrate this, we consider the IOI task again. According to \citet{wang2023ioi}, the S2 token (the position where one of the names occurs a second time) triggers the duplicate detection, and the S-Inhibition heads read from it to suppress that name at the final position.

A naive check for suppression is whether S2's credit for predicting S is negative. However, no other token in the template (the indirect object, the verb, the object) has an obvious reason to contribute to predicting S, so credit for S lands mostly on S1 and S2, and S2's credit may stay positive simply because of this copying. We therefore design a comparative experiment to separate the suppression mechanism from the copying mechanism. We do token attribution for both IOI and the matched ABC controls of Section~\ref{sec:eval_composition}, in which no name repeats. Let $\mathrm{X}\to\mathrm{Y}$ denote the credit to token X when the target is Y with a contrastive distractor per condition. For IOI we report $\mathrm{S1}\to\mathrm{S}$ and $\mathrm{S2}\to\mathrm{S}$, the credits to two occurrences of the same token in one prompt. For ABC we report $\mathrm{C}\to\mathrm{C}$, the credit a name gets for predicting itself with nothing to suppress it (pure copying), and $\mathrm{C}\to\mathrm{B}$, the credit the same slot gets for an unrelated name (position alone). We also report a normalized score $E = 1 - \tfrac{\mathrm{S2}\to\mathrm{S}\,-\,\mathrm{C}\to\mathrm{B}}{\mathrm{C}\to\mathrm{C}\,-\,\mathrm{C}\to\mathrm{B}}$, the fraction of copying credit removed at S2: $E \approx 1$ means suppression erased it, $E \approx 0$ that it survived, allowing comparison across methods.

Table~\ref{tab:suppression} shows the results. All methods capture some suppression signal: every $\mathrm{S2}\to\mathrm{S}$ value is smaller than the corresponding $\mathrm{C}\to\mathrm{C}$ and $\mathrm{S1}\to\mathrm{S}$. IFR shows no copying signal on the ABC prompts ($\mathrm{C}\to\mathrm{C} = \mathrm{C}\to\mathrm{B}$), so $E$ is undefined for it. Our method shows the strongest suppression: its $\mathrm{S2}\to\mathrm{S}$ falls below even its own positional baseline $\mathrm{C}\to\mathrm{B}$, and its $E$ exceeds that of AttnLRP and CP-LRP ($E > 1$: S2 receives less credit than the position earns for an arbitrary name).

Note that there is no ground truth for this comparison: the methods differ in their propagation rules, so a higher $E$ does not mean a better method, and we do not claim to beat any baseline in Table~\ref{tab:suppression}. This is precisely what the unified view adds. Because our credit follows from the two assumptions alone, the numbers have a definite reading: the $\mathrm{S2}\to\mathrm{S}$ column measures paths from S2 that, through the key--value interactions, end up suppressing the prediction of its own name. The same paths can be inspected directly (Section~\ref{sec:eval_composition}), and they name the mechanism: S-Inhibition heads reading S2 through their value branch.

Finally, the same mechanism appears across scales: the S1--S2 credit gap is positive in every Pythia model from 160M to 6.9B, with S2's credit turning outright negative at three of the five scales (Appendix~\ref{app:scaling}).

Taken together, the two assumptions appear to suffice: the single algorithm they fix clears an established reference for each readout, with interaction scores predicting knockout effects, rerooted paths recovering causally established compositions branch by branch, and token credit passing perturbation faithfulness. Where the readouts overlap they agree: the suppression mechanism the paths name is visible in the token credit. And where our method diverges from the causal ones, at layer 0, the divergence is itself interpretable as a prompt-independent bias.

\begin{table}
\centering
\caption{Copying vs.\ suppression at S2 (GPT-2 small, $n=100$).
Each condition uses a contrastive target (Appendix~\ref{app:edge_cases}):
S1$\to$S and S2$\to$S attribute toward $W_U[\mathrm{S}]-W_U[\mathrm{IO}]$,
C$\to$C toward $W_U[\mathrm{C}]-W_U[\mathrm{A}]$, and C$\to$B toward
$W_U[\mathrm{B}]-W_U[\mathrm{A}]$. Notation and $E$ are defined in
the text. For IFR, C$\to$C $=$ C$\to$B, so $E$ is undefined. Entries are percentages of the prompt's total positive credit toward the
target; $E$ is unitless.} 
\label{tab:suppression}
\begin{tabular}{lccccc}
\toprule
Method & C$\to$C & S2$\to$S & C$\to$B & S1$\to$S & $E$ \\
\midrule
IFR     &  +7.2 &  +4.8 &  +7.2 & +11.2 &   n/a \\
AttnLRP & +54.5 &  +2.1 &  -4.3 & +43.7 & +0.89 \\
CP-LRP  & +73.9 & +26.2 &  +2.5 & +51.9 & +0.67 \\
\midrule
\textsc{Unpack} & +23.9 &  +7.3 &  +8.7 & +22.2 & +1.09 \\
\bottomrule
\end{tabular}
\end{table}

\section{Limitations and Conclusion}
\label{sec:limitations}
\label{sec:conclusion}

We asked whether two structural assumptions, linearity of the residual stream and a shared key-value form, suffice for mechanistic interpretability. Our results indicate that they do for all three readouts studied here. A single computational graph provides an established external check for each: interaction scores predict knockout effects, rerooted paths recover causally established compositions branch by branch, and token credit passes a perturbation faithfulness test. The readouts are also connected: token credit decomposes into named paths, and the paths aggregate into component interactions. The systematic divergence from causal methods occurs at layer 0, where it is consistent with a prompt-independent bias. The resulting scores therefore have a definite meaning because they follow from stated assumptions that a reader can accept or reject, rather than from task-specific propagation rules.

Several choices and limitations remain. First, the decomposition is marginal rather than exact through LayerNorm: each component is centered, scaled by the LayerNorm weight, and divided by the full stream's standard deviation, which we treat as constant so that the decomposition remains additive. This treatment is motivated by the pre-norm architecture used by GPT-2 and Pythia; post-norm models require a different treatment and are not tested. The safe denominator and output-aligned projection are additional algorithmic choices, although the results are stable across the configurations examined in Appendix~\ref{app:beta}. Second, the method measures contribution under the computation the model actually performed, not causal effect under intervention. A large contribution need not imply that a component is causally necessary. Our layer-0 results illustrate this distinction: these components receive high credit across prompts, while the causally established components lose credit when the task is removed. Separating bias from signal required a matched counterfactual template, however, and the same diagnosis is unavailable when no such counterfactual exists, as in the greater-than task (Appendix~\ref{app:greater_than}). Contribution therefore complements intervention rather than replacing it. Whether it can predict interventional effect more generally, or identify a circuit without a counterfactual, remains open.

Finally, the method has computational limits. Its preparation cost is $O(L^2 H n)$ key-side scores for $L$ layers, $H$ heads, and sequence length $n$. Streaming one source layer at a time keeps memory linear rather than quadratic in $L$, but preparation time grows with model width and depth. Named-path enumeration is combinatorial in depth, so the pruning threshold $\tau$ and the top-$k$ paths retained per rerooting can miss low-magnitude routes in deeper models. The aggregate token attributions are computed exactly; the threshold affects only path enumeration (Appendix~\ref{app:pseudocode}).

This view places interaction, composition, and token attribution in one framework. Instead of developing a separate analysis for each question, we can study these aspects of the model through the same graph. This also gives us a clear direction for future work: use the graph to identify important computational routes, test them with interventions, and compare how they change across prompts and model scales. The graph could therefore serve as a common starting point for building and testing circuits.


\bibliography{iclr2027_conference}
\bibliographystyle{iclr2027_conference}


\appendix
\section{Algorithm details}
\label{app:algorithm}

Section~\ref{sec:recursive} states the two stages of the
recursion. This appendix gives the \textsc{SafeDenom} floor they
normalize with (Section~\ref{app:safedenom}), elaborates both
stages (Section~\ref{app:variants}), and assembles them as
pseudocode (Section~\ref{app:pseudocode}).

\subsection{The \textsc{SafeDenom} soft floor}
\label{app:safedenom}

The recursion repeatedly normalizes a parent's importance by a
denominator built from signed child contributions $\{r_i\}$:
\begin{equation*}
    \text{share}_i = \frac{r_i}{\sum_j r_j}.
\end{equation*}
When positive and negative contributions nearly cancel,
$|\sum_j r_j| \ll \sum_j |r_j|$, the per-share factor blows up
and amplifies the parent's credit. Over
deep recursion this compounds into unstable attributions where a
handful of paths receive scores far larger than the model's
actual logit can justify.

We control this with a soft floor on the denominator's magnitude:
\begin{equation}
    \textsc{SafeDenom}(\{r_i\}; \beta) :=
      \mathrm{sign}\!\Bigl(\sum_i r_i\Bigr) \cdot
      \max\!\Bigl(\bigl|\sum_i r_i\bigr|,\; \beta \cdot \sum_i |r_i|\Bigr).
    \label{eq:safe_denom}
\end{equation}
The maximum caps the per-step amplification at $1/\beta$. If the
signed sum is large in magnitude relative to the absolute mass
($|\sum_i r_i| \geq \beta \sum_i |r_i|$), \textsc{SafeDenom}
agrees with the signed sum and recovers the standard
normalization. If the signed sum is small (heavy cancellation),
the denominator's magnitude is held above $\beta$ times the
absolute mass while its sign is preserved.

We use $\beta = 0.8$ throughout; sensitivity to $\beta$ is
reported in Appendix~\ref{app:beta}.

\subsection{The two stages in full}
\label{app:variants}

\paragraph{Stage 1: credit to the entries.}
Stage 1 is Eq.~\ref{eq:stage1}, instantiated per sublayer. For a
head, the entries are source positions $s$, with
$\phi(S)_s = \alpha_{h,q,s}$ and $U_s = X_s W_V W_O$; for an MLP,
they are neurons $j$, with $\phi(S)_j = \phi(\text{pre}_j)$ and
$U_j = W_{\text{down}}[j,:]$. All vectors carry the marginal
LayerNorm normalization of Section~\ref{sec:limitations}.

At the first call $d = d_t$. Deeper in the recursion, $d$ is the
output the sublayer actually wrote at $q$:
\begin{equation}
    o_{\text{mlp}}(q) = \sum_j \phi(\text{pre}_j)\, W_{\text{down}}[j,:],
    \qquad
    o_{\text{attn}}(q) = \sum_{h' \in \text{layer}} \; \sum_s
        \alpha_{h',q,s}\, U^{(h')}_s .
    \label{eq:attn_aligned}
\end{equation}
$o_{\text{attn}}(q)$ is what the attention sublayer wrote at $q$, so
all heads of a layer are scored against it; the recursion expands one
head at a time and renormalizes that head's row with
\textsc{SafeDenom}. Appendix~\ref{app:configs} ablates this choice.

\paragraph{Stage 2 at a head: the three branches.}
Each source position's share $I_s$ is split three ways,
$M \in \{K, Q, V\}$, with weights $w_K = w_Q = w_V = \tfrac13$. The
three have the same form,
\begin{equation}
    I_M(c_j) = w_M\, I_s \cdot
        \frac{s^M_j}{\textsc{SafeDenom}(\{s^M_m\}_m,\, \beta)},
    \label{eq:branches}
\end{equation}
and differ only in the score $s^M_j$ they divide and the position
the branch continues at:
\begin{center}
\begin{tabular}{cllc}
\toprule
$M$ & score $s^M_j$ & upstream components & continues at \\
\midrule
K & $\langle X_q,\; c_{js} \rangle_{\text{QK}}$ & writers at $s$ & $s$ \\
Q & $\langle c_{jq},\; X_s \rangle_{\text{QK}}$ & writers at $q$ & $q$ \\
V & $\langle c_{js} W_V,\; X_s W_V \rangle$     & writers at $s$ & $s$ \\
\bottomrule
\end{tabular}
\end{center}
K and Q are the key-side and query-side splits of
Section~\ref{sec:framework}, the same bilinear form read on
its two arguments. V is a
component's share of the value the head actually formed at $s$, so
the V scores sum to $\lVert X_s W_V \rVert^2$; it is taken in head
space, before $W_O$, which is applied to the head's output as a whole
and scales every component's share alike. No credit direction enters
any of the three: each divides $I_s$ by the upstream components'
shares of a quantity the forward pass already computed.

\paragraph{Stage 2 at an MLP: one branch.}
The values $W_{\text{down}}[j,:]$ are parameters, so only the score
carries upstream credit. Neuron $j$'s share is distributed over the
components $c_{kq}$ writing the residual at $q$, by their
contribution to its pre-activation,
$s_k^{\text{mlp}}(q)[j] = c_{kq} W_{\text{up}}[:, j]$, normalized by
\textsc{SafeDenom} as above. Neurons are treated independently, and
credit flowing through different neurons recombines at each upstream
component. The recursion continues at $q$.

\subsection{Pseudocode}
\label{app:pseudocode}

Algorithm~\ref{alg:recursive} assembles the two stages. Each
component is passed down with the position it sits at: the K- and
V-branches continue at the source position, the Q-branch at the
query position.

\begin{algorithm}[h!]
\caption{Recursive backward attribution}
\label{alg:recursive}
\begin{algorithmic}[1]
\REQUIRE target token $t$, position $p$, pruning threshold $\tau$,
  amplification floor $\beta$ (Eq.~\ref{eq:safe_denom})
\STATE $d_t \leftarrow$ target direction (Eq.~\ref{eq:target_dir})
\STATE $I(c_k) \leftarrow \langle c_k(p),\, d_t \rangle$ for each
       residual component $c_k$ at position $p$
\STATE \textsc{Recurse}($\{(c_k,\, p,\, I(c_k))\}_k$, $d_t$)
\end{algorithmic}

\vspace{0.5em}
\noindent\textbf{Procedure} \textsc{Recurse}(items, $d$)
\begin{algorithmic}[1]
\FOR{each (component $c$, position $q$, credit $I$) in items with
     $|I| \geq \tau$}
  \IF{$c$ is the embedding}
    \STATE $\mathrm{credit}[q] \mathrel{+}= I$
  \ELSIF{$c$ is attention head $(L, h)$}
    \STATE $\{(s, I_s)\} \leftarrow \textsc{Stage1}(I, d)$, entries $=$ source positions
      \hfill (Eq.~\ref{eq:stage1})
    \FOR{each $s$ with $|I_s| \geq \tau$}
      \STATE \textsc{Recurse}($\{(c_j, s, I_K)\}_j \cup \{(c_j, q, I_Q)\}_j
             \cup \{(c_j, s, I_V)\}_j$, \; $o_{\text{attn}}(q)$)
        \hfill (Eq.~\ref{eq:branches})
    \ENDFOR
  \ELSIF{$c$ is MLP at layer $L$}
    \STATE $\{(j, I_j)\} \leftarrow \textsc{Stage1}(I, d)$, entries $=$ neurons
      \hfill (Eq.~\ref{eq:stage1})
    \FOR{each neuron $j$ with $|I_j| \geq \tau$}
      \STATE \textsc{Recurse}($\{(c_k,\, q,\, I_j(c_k))\}_k$, \; $o_{\text{mlp}}(q)$)
    \ENDFOR
  \ENDIF
\ENDFOR
\end{algorithmic}
\end{algorithm}

Here $o_{\text{attn}}(q)$ and $o_{\text{mlp}}(q)$ are the outputs the
head and the MLP actually wrote at $q$, the directions of
Eq.~\ref{eq:attn_aligned}. They are passed down only so that the next
call's Stage 1 has a direction to project on; Stage 2 never uses one.

The pruning threshold $\tau$ keeps the recursion tree tractable by
dropping sub-credits below $\tau$ in absolute value, which is
necessary because the tree expands by roughly $K \cdot S$ per level
(components times sequence length). The aggregate token-level credit
reported in Section~\ref{sec:evaluation} is computed by an
equivalent vectorized procedure that does not depend on $\tau$ and
so gives the exact attribution; only named-path enumeration uses
$\tau = 10^{-3}$.


\section{Edge cases}
\label{app:edge_cases}

This appendix records small implementation choices that affect
the numbers reported in the paper but are not part of the main
algorithm description.

\paragraph{Bias terms.}
Attention and MLP bias parameters are constants and so carry no
per-input information. We drop them from the residual
decomposition; the recursion does not visit bias components.

\paragraph{Logit-difference targets.}
For contrastive tasks such as IOI, we replace
$W_U[t] - \overline{W}_U$ in Eq.~\ref{eq:target_dir} with
$W_U[t] - W_U[t']$, where $t'$ is the distractor token (e.g.\
the subject S in IOI). Then $\langle c_k, d_t \rangle$ is what
$c_k$ contributes to the logit \emph{difference} $z_t - z_{t'}$
rather than to $z_t$ alone. Vocabulary-mean centering is
unnecessary in this case, since a constant added to every logit
cancels in the difference.

\paragraph{Pre-norm assumption.}
The marginal-LayerNorm normalization of
Section~\ref{sec:limitations} treats each component $c_k$ as
passing through one LayerNorm before being projected, which
matches the pre-norm architecture used by GPT-2 and the Pythia
family. Post-norm models (e.g.\ the original BERT) place
LayerNorm after the residual addition rather than before
projection, which would require a different decomposition.


\section{Knockout validation}
\label{app:knockout_full}

Section~\ref{sec:eval_knockout} states the setup and the
Pythia-6.9B result; this appendix gives the scoring formulas, the
sampling, and the full five-model correlations. All models are
taken at training step $143{,}000$.

\paragraph{Scores and ablation.}
The \emph{attention strength} of a component $k$ is its
interaction score of Eq.~\ref{eq:std} summed over all downstream
heads $m$ and query positions $q$,
$\sum_{m,q}\text{score}_{k\to m}^{\text{attn}}$; the \emph{MLP
strength} is the analogous sum of Eq.~\ref{eq:mlp} over downstream
MLPs, $\sum_{m,q}\text{score}_{k\to m}^{\text{mlp}}$. The
$\Delta\text{ppl}$ of the ablation of
Section~\ref{sec:eval_knockout} is measured on $200$ held-out Pile
sentences ($7{,}967$ tokens). For tractability
we select, per source layer, $5$ components spread across the range of
observed scores, yielding $56$, $116$, $116$, $156$, and $156$
knockouts at the $160$M, $410$M, $1.4$B, $2.8$B, and $6.9$B scales
respectively.

\begin{figure}[h]
\centering
\begin{subfigure}[b]{0.72\textwidth}
  \centering
  \includegraphics[width=\textwidth]{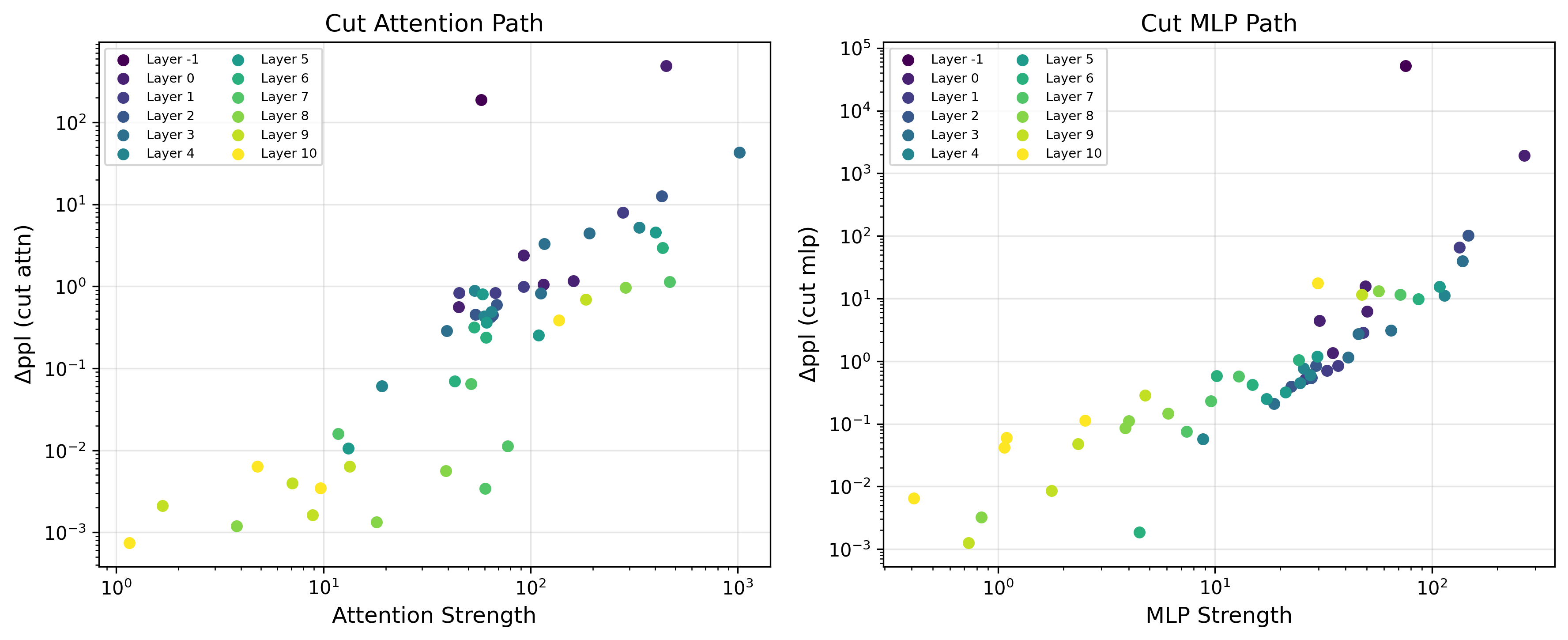}
  \caption{Pythia-160M-deduped (12 layers, 56 knockouts).}
  \label{fig:knockout_160m}
\end{subfigure}\\[0.2em]
\begin{subfigure}[b]{0.72\textwidth}
  \centering
  \includegraphics[width=\textwidth]{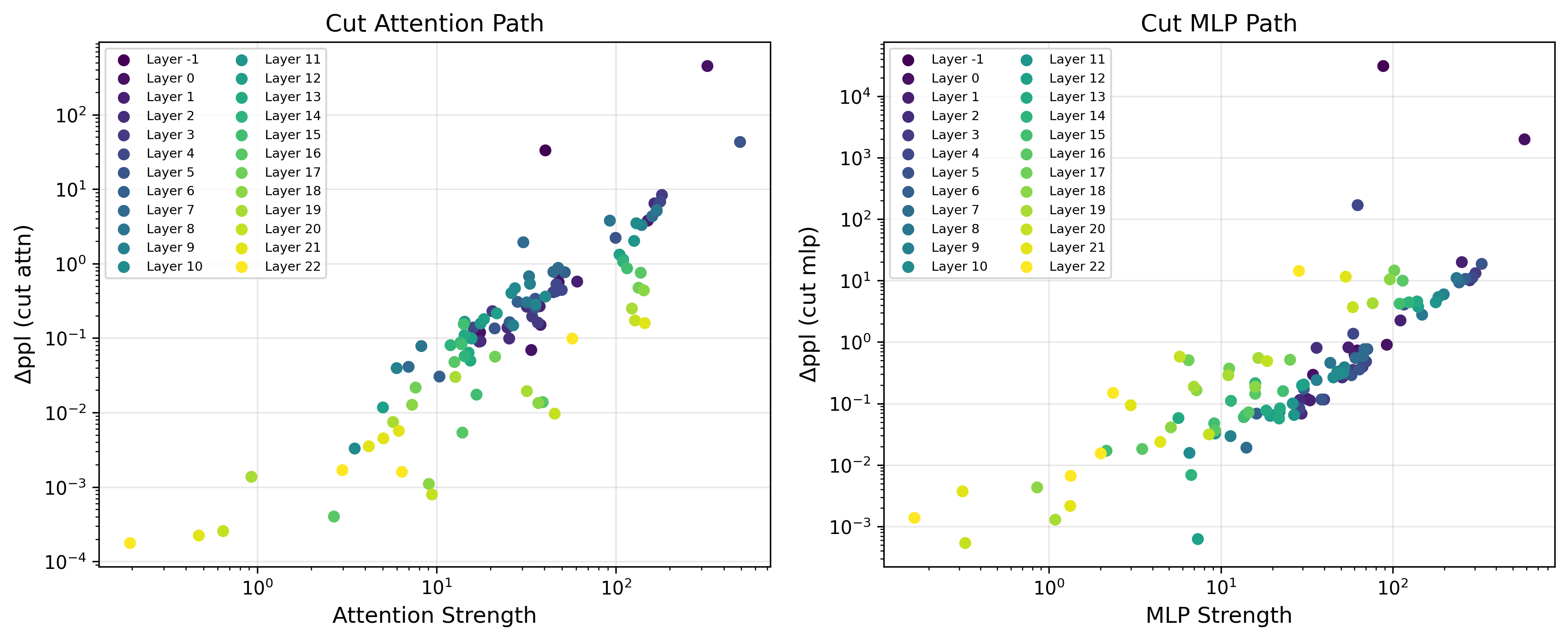}
  \caption{Pythia-410M-deduped (24 layers, 116 knockouts).}
  \label{fig:knockout_410m}
\end{subfigure}\\[0.2em]
\begin{subfigure}[b]{0.72\textwidth}
  \centering
  \includegraphics[width=\textwidth]{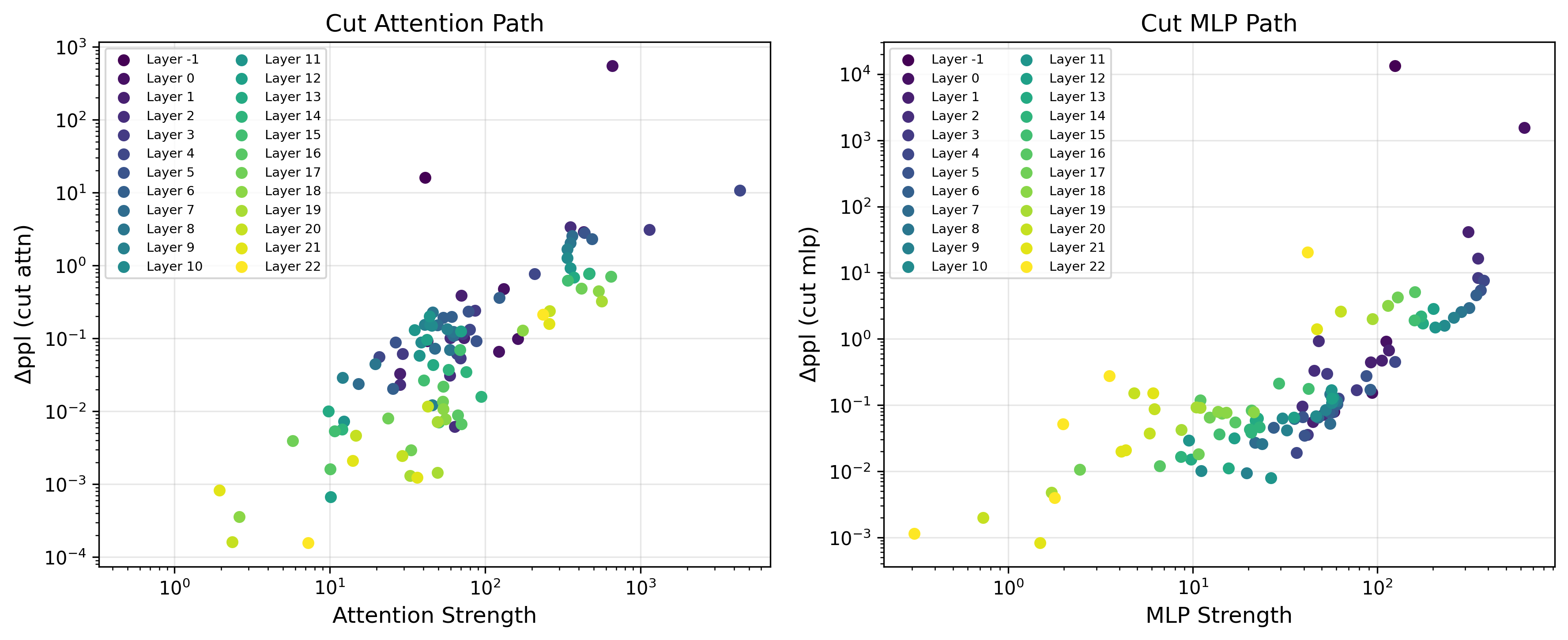}
  \caption{Pythia-1.4B-deduped (24 layers, 116 knockouts).}
  \label{fig:knockout_1_4b}
\end{subfigure}\\[0.2em]
\begin{subfigure}[b]{0.72\textwidth}
  \centering
  \includegraphics[width=\textwidth]{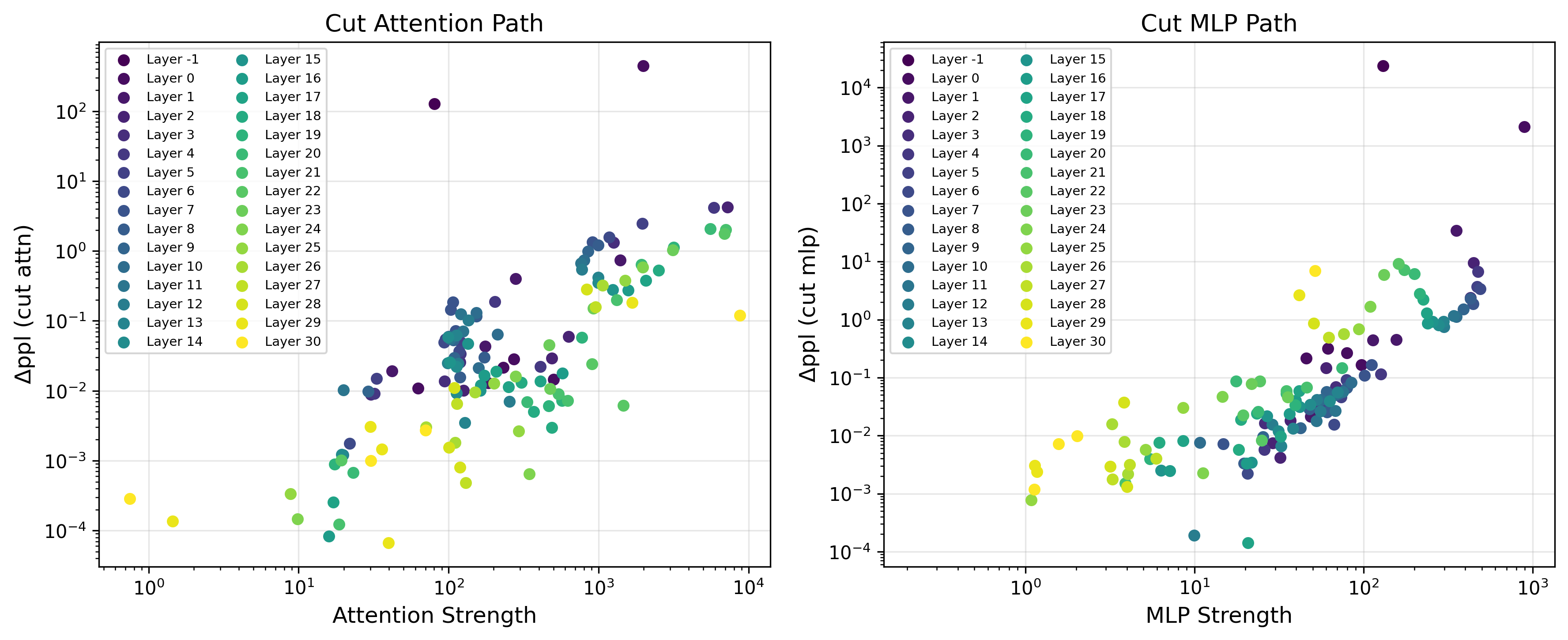}
  \caption{Pythia-2.8B-deduped (32 layers, 156 knockouts).}
  \label{fig:knockout_2_8b}
\end{subfigure}
\caption{Knockout validation across the Pythia-deduped family;
Pythia-6.9B is shown in Figure~\ref{fig:knockout_6_9b}. Each point
is one residual-stream component: the $x$-axis its aggregated
interaction score, the $y$-axis $\Delta\text{ppl}$ when the
channel is ablated (left panels: attention; right panels: MLP).
Points are colored by source layer (darker $=$ earlier; layer $-1$
is the embedding); both axes log-scaled. The monotonic relationship
is stable across the $40\times$ parameter range.}
\label{fig:knockout_all}
\end{figure}

\paragraph{Results.}
Figure~\ref{fig:knockout_all} plots score against knockout effect
for each of the five models; the monotonic relationship of
Figure~\ref{fig:knockout_6_9b} holds on both channels at every
scale. Table~\ref{tab:knockout_rho} reports Spearman rank
correlations: within-layer mean $\rho$ between $0.86$ and $0.96$ on the
MLP channel and between $0.72$ and $0.86$ on the attention channel.
The MLP channel is cleanly ranked at every scale; the attention channel
is noisier but still strong, reflecting that attention strength predicts
a more indirect outcome (a perturbation to downstream attention
patterns) than MLP strength does (a direct perturbation to downstream
MLP activations). Two qualitative patterns appear in every model: the
embedding (layer $-1$) is a high-end outlier on the MLP-cut channel,
with $\Delta\text{ppl} > 10^3$ at every scale; and layer identity is
visible as a colour gradient from purple (early) to yellow (late), with
early layers dominating the MLP channel and late-layer attention heads
dominating the attention channel.

\begin{table}[h]
\centering
\caption{Spearman rank correlation between interaction score and
knockout $\Delta\text{ppl}$. \emph{Within-layer} $\rho$ is the
mean of per-source-layer Spearman correlations (the stricter
test, holding the set of downstream receivers fixed).
\emph{Cross-layer} $\rho$ pools all components for a given
channel.}
\label{tab:knockout_rho}
\begin{tabular}{lcccccc}
\toprule
 & & \multicolumn{2}{c}{within-layer $\rho$} & \multicolumn{2}{c}{cross-layer $\rho$} \\
\cmidrule(lr){3-4} \cmidrule(lr){5-6}
Model & $n$ & attn & mlp & attn & mlp \\
\midrule
Pythia-160M & 56  & 0.72 & 0.96 & 0.84 & 0.95 \\
Pythia-410M & 116 & 0.82 & 0.91 & 0.88 & 0.87 \\
Pythia-1.4B & 116 & 0.73 & 0.86 & 0.78 & 0.82 \\
Pythia-2.8B & 156 & 0.73 & 0.87 & 0.67 & 0.86 \\
Pythia-6.9B & 156 & 0.86 & 0.90 & 0.76 & 0.87 \\
\bottomrule
\end{tabular}
\end{table}

\section{Composition ranking matrices}
\label{app:composition_matrices}

Table~\ref{tab:comp-matrices} reports composition rankings for the
first upstream head from each role tier, using attention heads from
layer~1 onward. Each cell gives the median rank and percentage of
rerootings in which the upstream role was found; lower rank indicates
stronger composition. Figure~\ref{fig:composition-paths} illustrates
the route-level view for the top-3 V-mode paths from a rerooting at
S-Inhibition head A8.H6.

\begin{table}[h]
\centering
\caption{Composition ranking matrices ($n{=}100$, GPT-2 small). Each cell:
median rank / found\%. Rows = head of interest, columns = upstream role.
Lower rank indicates stronger composition.}
\label{tab:comp-matrices}
\small
\setlength{\tabcolsep}{5pt}
\begin{tabular}{ll ccccc}
\toprule
Filter & $\downarrow$ / $\uparrow$ & NM & S-Inh & Ind & Dup & Prev \\
\midrule
\multirow{3}{*}{All modes}
 & NM    & 6 / 33\%  & 4 / 100\% & 14 / 84\% & 22 / 29\% & 21 / 77\% \\
 & S-Inh & --         & 20 / 32\% & 4 / 96\%  & 2 / 97\%  & 20 / 52\% \\
 & Ind   & --         & --         & 3 / 25\%  & 2 / 100\% & 4 / 98\% \\
\midrule
\multirow{3}{*}{K-mode}
 & NM    & --  & 10 / 48\% & 14 / 27\% & 13 / 19\% & 14 / 46\% \\
 & S-Inh & --  & --         & 5 / 93\%  & 4 / 91\%  & 13 / 4\%  \\
 & Ind   & --  & --         & 21 / 3\%  & 8 / 10\%  & 3 / 98\% \\
\midrule
\multirow{3}{*}{Q-mode}
 & NM    & 2 / 33\%  & 1 / 100\% & 4 / 63\%  & 18 / 5\%  & 10 / 56\% \\
 & S-Inh & --         & 9 / 32\%  & 3 / 66\%  & 13 / 35\% & 11 / 42\% \\
 & Ind   & --         & --         & 2 / 25\%  & 1 / 100\% & 10 / 36\% \\
\midrule
\multirow{3}{*}{V-mode}
 & NM    & -- & --         & 4 / 24\%  & 6 / 10\%  & 6 / 9\%   \\
 & S-Inh & -- & --         & 2 / 96\%  & 1 / 95\%  & 10 / 21\% \\
 & Ind   & -- & --         & 8 / 10\%  & 11 / 32\% & 9 / 88\%  \\
\bottomrule
\end{tabular}
\end{table}

\begin{figure}[h]
\centering
\resizebox{0.75\textwidth}{!}{
\definecolor{pc0}{RGB}{74,127,181}
\definecolor{pc1}{RGB}{212,128,78}
\definecolor{pc2}{RGB}{106,171,115}
\begin{tikzpicture}[x=1cm, y=1cm]

  \node[ll] at (4.80,2.60) {L0};
  \draw[ruleG, line width=0.25pt] (4.20,2.60) -- (12.40,2.60);
  \node[ll] at (4.80,5.00) {L5};
  \draw[ruleG, line width=0.25pt] (4.20,5.00) -- (12.40,5.00);
  \node[ll] at (4.80,6.60) {L6};
  \draw[ruleG, line width=0.25pt] (4.20,6.60) -- (12.40,6.60);
  \node[ll] at (4.80,8.20) {L8};
  \draw[ruleG, line width=0.25pt] (4.20,8.20) -- (12.40,8.20);
  \node[ll] at (4.80,1.00) {emb};
  \draw[ruleG, line width=0.25pt] (4.20,1.00) -- (12.40,1.00);

  \node[tl] at (7.10,0.47) {S2};
  \node[pl] at (7.10,0.19) {pos 9};
  \node[text=labG] at (9.00,0.47) {$\cdots$};
  \node[tl] at (10.90,0.47) {END};
  \node[pl] at (10.90,0.19) {pos 13};

  \node[cs] at (9.530,8.150) {};
  \node[cs] at (5.730,3.350) {};
  \node[cs] at (7.130,0.950) {};
  \node[cs] at (5.730,4.950) {};
  \node[cs] at (5.730,6.550) {};

  \node[atn] (attn8head613) at (9.500,8.200) {A8.H6};
  \node[mlp] (mlp09) at (5.700,3.400) {MLP 0};
  \node[emb] (embedding9) at (7.100,1.000) {embed};
  \node[atn] (attn5head59) at (5.700,5.000) {A5.H5};
  \node[atn] (attn6head99) at (5.700,6.600) {A6.H9};
  \node[gho] (attn8head69ghost) at (5.700,8.200) {A8.H6};

  \begin{scope}[pc0, line width=1.3pt, opacity=0.80]
    \draw (6.960,1.310) -- (6.960,3.020);
    \draw[ar] (6.960,3.020) .. controls (6.960,3.010) and (5.560,2.940) .. (5.560,3.090);
    \draw (5.560,3.710) .. controls (5.560,3.860) and (6.960,3.860) .. (6.960,3.780);
    \draw (6.960,3.780) -- (6.960,7.820);
    \draw[ar] (6.960,7.820) .. controls (6.960,7.810) and (5.560,7.740) .. (5.560,7.890);
    \draw[dar] (6.600,8.060) -- (8.600,8.060);
    \node[mt, text=pc0] at (7.600,8.490) {V};
  \end{scope}
  \begin{scope}[pc1, line width=1.3pt, opacity=0.35]
    \draw (7.100,1.310) -- (7.100,3.020);
    \draw[ar] (7.100,3.020) .. controls (7.100,3.010) and (5.700,2.940) .. (5.700,3.090);
    \draw (5.700,3.710) .. controls (5.700,3.860) and (7.100,3.860) .. (7.100,3.780);
    \draw (7.100,3.780) -- (7.100,4.620);
    \draw[ar] (7.100,4.620) .. controls (7.100,4.610) and (5.700,4.540) .. (5.700,4.690);
    \node[mt, anchor=east, text=pc1] at (4.700,4.770) {Q};
    \draw (5.700,5.310) .. controls (5.700,5.460) and (7.100,5.460) .. (7.100,5.380);
    \draw (7.100,5.380) -- (7.100,7.820);
    \draw[ar] (7.100,7.820) .. controls (7.100,7.810) and (5.700,7.740) .. (5.700,7.890);
    \draw[dar] (6.600,8.200) -- (8.600,8.200);
  \end{scope}
  \begin{scope}[pc2, line width=1.3pt, opacity=0.35]
    \draw (7.240,1.310) -- (7.240,3.020);
    \draw[ar] (7.240,3.020) .. controls (7.240,3.010) and (5.840,2.940) .. (5.840,3.090);
    \draw (5.840,3.710) .. controls (5.840,3.860) and (7.240,3.860) .. (7.240,3.780);
    \draw (7.240,3.780) -- (7.240,6.220);
    \draw[ar] (7.240,6.220) .. controls (7.240,6.210) and (5.840,6.140) .. (5.840,6.290);
    \node[mt, anchor=east, text=pc2] at (4.700,6.370) {Q};
    \draw (5.840,6.910) .. controls (5.840,7.060) and (7.240,7.060) .. (7.240,6.980);
    \draw (7.240,6.980) -- (7.240,7.820);
    \draw[ar] (7.240,7.820) .. controls (7.240,7.810) and (5.840,7.740) .. (5.840,7.890);
    \draw[dar] (6.600,8.340) -- (8.600,8.340);
  \end{scope}

  \fill[white, rounded corners=4pt, draw=ruleG, line width=0.3pt] (4.20,9.40) rectangle (12.40,10.90);
  \draw[pc0, line width=2pt, line cap=round] (4.60,10.55) -- ++(0.35,0);
  \node[font=\sffamily\tiny, text=tokC, anchor=west] at (5.10,10.55) {emb@S2 $\to$ MLP0@S2 $\xrightarrow{\text{V}}$ A8.H6@END \enspace +14.5\%};
  \draw[pc1, line width=2pt, line cap=round] (4.60,10.15) -- ++(0.35,0);
  \node[font=\sffamily\tiny, text=tokC, anchor=west] at (5.10,10.15) {emb@S2 $\to$ MLP0@S2 $\xrightarrow{\text{Q}}$ \textbf{Ind} (A5.H5)@S2 $\xrightarrow{\text{V}}$ A8.H6@END \enspace +2.5\%};
  \draw[pc2, line width=2pt, line cap=round] (4.60,9.75) -- ++(0.35,0);
\node[font=\sffamily\tiny, text=tokC, anchor=west] at (5.10,9.75) {emb@S2 $\to$ MLP0@S2 $\xrightarrow{\text{Q}}$ \textbf{Ind} (A6.H9)@S2 $\xrightarrow{\text{V}}$ A8.H6@END \enspace +1.5\%};

\end{tikzpicture}}
\caption{Top-3 V-mode composition paths from a rerooting at S-Inhibition
head A8.H6. The paths trace upstream computation through cross-position
attention; the second and third paths pass through Induction heads
A5.H5 and A6.H9, illustrating the Ind$\to$S-Inh composition. Percentages are shares of the total positive credit entering the root \texttt{A8.H6@END}.}
\label{fig:composition-paths}
\end{figure}

Table~\ref{tab:comp-matrices-l0} repeats the rankings with layer-0
heads included. This mainly introduces two Duplicate Token heads,
\texttt{attn\_0\_head\_1} and \texttt{attn\_0\_head\_10}, which move the
Dup column to ranks 1--2 with near-100\% coverage. Because these heads
write at every position, the same pattern also appears on the ABC
counterfactual and therefore does not provide additional evidence for
task-specific Dup composition. Other columns retain essentially the
same coverage while their median ranks increase by about 10 as the
pool gains competitors.

\begin{table}[h]
\centering
\small
\setlength{\tabcolsep}{5pt}
\begin{tabular}{ll ccccc}
\toprule
Filter & $\downarrow$ / $\uparrow$ & NM & S-Inh & Ind & Dup & Prev \\
\midrule
\multirow{3}{*}{All modes}
 & NM    & 18 / 33\% & 15 / 100\% & 26 / 84\% & 2 / 100\% & 33 / 77\% \\
 & S-Inh & -- & 32 / 32\% & 12 / 96\% & 1 / 100\% & 31 / 52\% \\
 & Ind   & -- & -- & 13 / 25\% & 1 / 100\% & 12 / 98\% \\
\midrule
\multirow{3}{*}{K-mode}
 & NM    & -- & 21 / 48\% & 25 / 27\% & 5 / 100\% & 24 / 46\% \\
 & S-Inh & -- & -- & 13 / 93\% & 2 / 100\% & 23 / 4\% \\
 & Ind   & -- & -- & 32 / 3\% & 2 / 100\% & 7 / 98\% \\
\midrule
\multirow{3}{*}{Q-mode}
 & NM    & 6 / 33\% & 3 / 100\% & 9 / 63\% & 6 / 89\% & 18 / 56\% \\
 & S-Inh & -- & 18 / 32\% & 10 / 66\% & 3 / 100\% & 19 / 42\% \\
 & Ind   & -- & -- & 5 / 25\% & 1 / 100\% & 20 / 36\% \\
\midrule
\multirow{3}{*}{V-mode}
 & NM    & -- & -- & 14 / 24\% & 1 / 100\% & 16 / 9\% \\
 & S-Inh & -- & -- & 5 / 96\% & 1 / 100\% & 17 / 21\% \\
 & Ind   & -- & -- & 19 / 10\% & 2 / 100\% & 18 / 88\% \\
\bottomrule
\end{tabular}
\caption{Composition rankings with layer-0 attention heads included.
Compare Table~\ref{tab:comp-matrices} for the layer-${\geq}1$ pool.}
\label{tab:comp-matrices-l0}
\end{table}
\section{Coverage-drop details}
\label{app:coverage}

Table~\ref{tab:abc-drop-hyp-split} gives a per-head view of the
coverage drop, complementary to Figure~\ref{fig:coverage_drop}:
the figure counts appearances through the claimed branch only,
on the prompts where the model fails the ABC control; the table
sums credit over the K, Q and V branches on all 100 prompt pairs. For each hypothesis of Table~\ref{tab:comp_ioi} it lists the five strongest layer-${\geq}1$ and five strongest layer-0 heads on a single ranking over all attention heads, re-rootings paired across arms. Under both views the predicted tiers lose share and coverage on the counterfactual while layer-0 heads hold near-100\% coverage on both arms, the bias behavior of Appendix~\ref{app:composition_matrices}.

\begin{table}[h]
\centering
\caption{Per-head coverage and credit share for the three composition
hypotheses ($n{=}100$, GPT-2 small). The five strongest heads are shown
for each layer split. IOI and ABC rerootings are paired and use a
single ranking over all heads. Share is mean credit fraction, with
absences counted as zero; Cov.\ is the fraction of rerootings in which
the head appears; $r_{\mathrm{ABC}}$ is its ABC rank. Layer-0 heads
retain near-100\% coverage on both arms and therefore do not distinguish
the task from its counterfactual.}
\label{tab:abc-drop-hyp-split}
\small
\setlength{\tabcolsep}{5pt}
\begin{tabular}{l c S[table-format=2.2] S[table-format=2.2] r S[table-format=3.1] S[table-format=3.1] r r}
\toprule
 & & \multicolumn{3}{c}{Credit share \%} & \multicolumn{3}{c}{Coverage \%} & \\
\cmidrule(lr){3-5} \cmidrule(lr){6-8}
Head & Tier & {IOI} & {ABC} & {rel.} & {IOI} & {ABC} & {$\Delta$} & {$r_{\mathrm{ABC}}$} \\
\midrule
\multicolumn{9}{l}{\textbf{H1} \quad NM $\leftarrow$ S-Inh \quad \textit{root: name movers, END; predicts S-Inh}} \\
\addlinespace[2pt]
\multicolumn{9}{l}{\quad\itshape layer${\geq}1$} \\
\quad \texttt{attn\_1\_head\_4} & -- & 2.26 & 1.53 & -32\% & 94.3 & 83.3 & -11.0 & 14 \\
\quad \texttt{attn\_1\_head\_8} & -- & 2.24 & 1.94 & -13\% & 90.7 & 95.7 & +5.0 & 13 \\
\quad \texttt{attn\_1\_head\_1} & -- & 1.90 & 1.17 & -38\% & 96.3 & 92.0 & -4.3 & 15 \\
\quad \texttt{attn\_8\_head\_6} & S-Inh & 1.46 & 0.60 & -59\% & 100.0 & 91.3 & -8.7 & 17 \\
\quad \texttt{attn\_1\_head\_10} & -- & 1.40 & 1.06 & -24\% & 100.0 & 98.0 & -2.0 & 16 \\
\addlinespace[2pt]
\multicolumn{9}{l}{\quad\itshape layer 0} \\
\quad \texttt{attn\_0\_head\_8} & -- & 13.26 & 13.61 & +3\% & 100.0 & 100.0 & +0.0 & 2 \\
\quad \texttt{attn\_0\_head\_1} & Dup & 13.09 & 16.71 & +28\% & 100.0 & 100.0 & +0.0 & 1 \\
\quad \texttt{attn\_0\_head\_9} & -- & 9.77 & 9.11 & -7\% & 100.0 & 100.0 & +0.0 & 5 \\
\quad \texttt{attn\_0\_head\_4} & -- & 9.32 & 10.73 & +15\% & 100.0 & 100.0 & +0.0 & 4 \\
\quad \texttt{attn\_0\_head\_5} & -- & 9.25 & 11.32 & +22\% & 100.0 & 100.0 & +0.0 & 3 \\
\midrule
\multicolumn{9}{l}{\textbf{H2} \quad S-Inh $\leftarrow$ Dup/Ind \quad \textit{root: S-Inh heads, END; predicts Dup/Ind}} \\
\addlinespace[2pt]
\multicolumn{9}{l}{\quad\itshape layer${\geq}1$} \\
\quad \texttt{attn\_1\_head\_8} & -- & 6.15 & 7.71 & +25\% & 99.5 & 100.0 & +0.5 & 6 \\
\quad \texttt{attn\_3\_head\_0} & Dup & 4.39 & 0.21 & -95\% & 96.5 & 55.5 & -41.0 & 27 \\
\quad \texttt{attn\_1\_head\_4} & -- & 3.34 & 4.87 & +46\% & 100.0 & 100.0 & +0.0 & 8 \\
\quad \texttt{attn\_5\_head\_5} & Ind & 3.01 & 0.25 & -92\% & 95.0 & 57.2 & -37.8 & 25 \\
\quad \texttt{attn\_1\_head\_1} & -- & 1.25 & 1.41 & +12\% & 92.2 & 94.0 & +1.8 & 14 \\
\addlinespace[2pt]
\multicolumn{9}{l}{\quad\itshape layer 0} \\
\quad \texttt{attn\_0\_head\_11} & -- & 13.86 & 14.40 & +4\% & 100.0 & 100.0 & +0.0 & 2 \\
\quad \texttt{attn\_0\_head\_10} & Dup & 12.09 & 9.08 & -25\% & 100.0 & 100.0 & +0.0 & 4 \\
\quad \texttt{attn\_0\_head\_1} & Dup & 10.82 & 14.65 & +35\% & 100.0 & 100.0 & +0.0 & 1 \\
\quad \texttt{attn\_0\_head\_3} & -- & 8.80 & 10.44 & +19\% & 100.0 & 100.0 & +0.0 & 3 \\
\quad \texttt{attn\_0\_head\_4} & -- & 7.60 & 9.00 & +18\% & 100.0 & 100.0 & +0.0 & 5 \\
\midrule
\multicolumn{9}{l}{\textbf{H3} \quad Ind $\leftarrow$ Prev \quad \textit{root: induction heads, S2; predicts Prev/Dup}} \\
\addlinespace[2pt]
\multicolumn{9}{l}{\quad\itshape layer${\geq}1$} \\
\quad \texttt{attn\_3\_head\_0} & Dup & 5.58 & 0.47 & -92\% & 100.0 & 61.3 & -38.7 & 32 \\
\quad \texttt{attn\_1\_head\_8} & -- & 3.33 & 2.97 & -11\% & 98.8 & 99.8 & +1.0 & 12 \\
\quad \texttt{attn\_1\_head\_9} & -- & 3.29 & 6.47 & +97\% & 99.8 & 100.0 & +0.2 & 4 \\
\quad \texttt{attn\_1\_head\_4} & -- & 3.25 & 7.40 & +127\% & 100.0 & 100.0 & +0.0 & 3 \\
\quad \texttt{attn\_4\_head\_11} & Prev & 2.92 & 0.46 & -84\% & 97.2 & 80.5 & -16.8 & 33 \\
\addlinespace[2pt]
\multicolumn{9}{l}{\quad\itshape layer 0} \\
\quad \texttt{attn\_0\_head\_1} & Dup & 13.12 & 10.27 & -22\% & 100.0 & 100.0 & +0.0 & 1 \\
\quad \texttt{attn\_0\_head\_9} & -- & 8.30 & 10.00 & +21\% & 100.0 & 100.0 & +0.0 & 2 \\
\quad \texttt{attn\_0\_head\_11} & -- & 6.59 & 5.26 & -20\% & 100.0 & 100.0 & +0.0 & 8 \\
\quad \texttt{attn\_0\_head\_10} & Dup & 5.50 & 5.22 & -5\% & 100.0 & 100.0 & +0.0 & 9 \\
\quad \texttt{attn\_0\_head\_5} & -- & 5.25 & 6.17 & +17\% & 100.0 & 100.0 & +0.0 & 5 \\
\bottomrule
\end{tabular}
\end{table}

\section{Composition on the greater-than circuit}
\label{app:greater_than}

\paragraph{Setup.}
\label{app:gt_setup}
Prompts and target follow the released code of
\citet{hanna2023greater}. The template is ``\texttt{The \{noun\}
lasted from the year XXYY to the year XX}'', with their 120
FrameNet nouns and the 768 two-token years in $[1000,1900)$.
Single-token years and each century's first and last valid year
are excluded, so every prompt has at least one valid and one
invalid two-digit answer; YY is swept balanced over the cutoff,
and we do not filter for whether GPT-2 answers correctly. Their
metric is a group quantity, the probability difference
$\sum_{y>\mathrm{YY}} p_y - \sum_{y\le\mathrm{YY}} p_y$, so we
attribute toward the difference of the mean unembedding rows of
the valid and the invalid year sets, a logit-difference target in
the sense of Appendix~\ref{app:edge_cases}.

\paragraph{Claims and procedure.}
Their \S3.2 builds the circuit by iterated path patching and
states a set of composition claims, each naming what a given
receiver relies on. We check the four whose named sources sit
above layer $0$, since a layer-$0$ component contributes strongly
to every later layer and ranks high whatever it feeds:
\begin{enumerate}\itemsep1pt \parskip0pt \topsep3pt
\item MLP 9 relies on a9.h1.
\item MLP 8 relies on a8.h11, a8.h8, a7.h10, a6.h9, a5.h5 and a5.h1.
\item MLPs $8$--$11$ rely on the MLPs upstream of them.
\item Those seven heads read their values, at the YY position, from MLPs $1$--$3$.
\end{enumerate}
For each claim we reroot at the receivers it names, at the end
position for claims 1 to 3 and at the YY position for claim 4, and
rank the components feeding each receiver by the credit they send
in. A claim names a group of components rather than a single one,
so we score it by the best-placed member of the group, as
Section~\ref{sec:eval_composition} does for the role tiers on IOI;
ranks are medians over $n=100$ prompts. Each rank is reported
twice, against all upstream components and with the layer-$0$
heads, $\mathrm{mlp\_0}$ and the embedding taken out of the pool;
without the second, layer-$0$ components fill the top of nearly
every ranking and there is little to compare.

\begin{table}[h]
\centering
\caption{Composition claims 1--4 of \citet{hanna2023greater} ($n=100$, GPT-2 small). Each entry is
the median rank of the best-placed member of the group they name, in the receiver's upstream
ranking, against all upstream components and with layer-$0$ writers removed. \emph{recv.}\ is how
many receivers the claim covers; \emph{pool} is how many components the ranking runs over, so rank
$2$ of $17$ and rank $2$ of $60$ can be told apart.}
\label{tab:gt-claims}
\small
\setlength{\tabcolsep}{5pt}
\begin{tabular}{cl c cc cc}
\toprule
& & & \multicolumn{2}{c}{median rank} & \multicolumn{2}{c}{pool size} \\
\cmidrule(lr){4-5}\cmidrule(lr){6-7}
\# & claim & recv. & all & layer\,$\ge 1$ & all & layer\,$\ge 1$ \\
\midrule
1 & MLP 9 relies on a9.h1                        & 1 & 9.5 & 3.5 & 69     & 55     \\
2 & MLP 8 relies on six heads at layers $5$--$8$ & 1 & 10  & 2   & 74     & 60     \\
3 & MLPs $8$--$11$ rely on the MLPs upstream     & 3 & 4   & 2   & 60--69 & 47--55 \\
4 & the seven heads read values from MLPs $1$--$3$ & 7 & 9 & 2   & 31--39 & 17--23 \\
\bottomrule
\end{tabular}
\end{table}

All four claims place the group they name in the top four once
layer-$0$ writers are out of the pool (Table~\ref{tab:gt-claims}),
against pools of $17$ to $60$ components. Claim 3 is the cleanest:
$\mathrm{mlp\_8}$ is the single largest contributor to
$\mathrm{mlp\_9}$ in either pool. Against all sources the same
claims sit near rank $10$, and for most of them that gap is
layer-$0$ crowding rather than a weaker edge: everything ranked
above the group into $\mathrm{mlp\_8}$, and above the value inputs
of a9.h1 and a6.h9, is an embedding or a layer-$0$ writer. a5.h1
is the exception, with layer-$1$ and layer-$2$ heads above its
value inputs as well.

\begin{table}[h]
\centering
\caption{Per-source detail for claims 2, 3 and 4, as \emph{all\,/\,layer\,$\ge 1$}. A group rank
reports its best member; the spread behind it is wide, and two named sources
($\mathrm{a8.h8}\!\to\!\mathrm{mlp\_8}$, $\mathrm{mlp\_8}\!\to\!\mathrm{mlp\_11}$) stay low in
both pools.}
\label{tab:gt-detail}
\small
\setlength{\tabcolsep}{5pt}
\begin{tabular}{ll ll ll}
\toprule
\multicolumn{2}{c}{claim 2: into $\mathrm{mlp\_8}$} &
\multicolumn{2}{c}{claim 3: MLP to MLP} &
\multicolumn{2}{c}{claim 4: best of MLPs $1$--$3$} \\
\cmidrule(lr){1-2}\cmidrule(lr){3-4}\cmidrule(lr){5-6}
source & rank & edge & rank & receiver & rank \\
\midrule
a5.h5  & 10 / 2  & $8\!\to\!9$   & 1 / 1   & a9.h1  & 5 / 1   \\
a7.h10 & 16 / 7  & $8\!\to\!10$  & 4 / 2   & a7.h10 & 6 / 2   \\
a6.h9  & 21 / 10 & $9\!\to\!10$  & 5 / 2   & a8.h11 & 8 / 2   \\
a8.h11 & 29 / 17 & $10\!\to\!11$ & 10 / 3  & a8.h8  & 9 / 3   \\
a5.h1  & 29 / 18 & $9\!\to\!11$  & 15 / 4  & a5.h5  & 9.5 / 3 \\
a8.h8  & 44 / 31 & $8\!\to\!11$  & 39 / 27 & a6.h9  & 10 / 2  \\
       &         &               &         & a5.h1  & 26 / 14 \\
\bottomrule
\end{tabular}
\end{table}

Table~\ref{tab:gt-detail} gives the per-source detail behind the
group ranks. The spread inside a group is wide: of the six heads
feeding $\mathrm{mlp\_8}$, a5.h5 reaches rank $2$ while a8.h8 sits
at $31$. Claim 3 holds through the adjacent hops and weakens with
distance, from rank $1$ for $\mathrm{mlp\_8}\!\to\!\mathrm{mlp\_9}$
to $27$ for $\mathrm{mlp\_8}\!\to\!\mathrm{mlp\_11}$. On claim 4
every head but a5.h1 finds one of MLPs $1$--$3$ in its top three,
matching the joint form of the claim.

\begin{table}[h]
\centering
\caption{Circuit components against everything else, ranked in the same pool. Median rank over
every component appearing in any rerooting, split by whether \citet{hanna2023greater} name it
anywhere in the circuit. Unlike the per-claim ranks above, this comparison is not fixed by
construction, since both groups are ranked against the same competitors.}
\label{tab:gt-aggregate}
\small
\begin{tabular}{lcc}
\toprule
pool & circuit & non-circuit \\
\midrule
all sources    & 10 \; ($n=246$) & 22 \; ($n=661$) \\
layer\,$\ge 1$ & \phantom{0}8 \; ($n=121$) & 14 \; ($n=498$) \\
\bottomrule
\end{tabular}
\end{table}

Circuit components outrank the rest beyond the named connections
(Table~\ref{tab:gt-aggregate}): median rank $10$ against $22$ over
all sources, and $8$ against $14$ at layer $\ge 1$. That the gap
survives removing layer $0$ rules out the reading that circuit
components rank high only because early writers do and the circuit
contains early writers. Which of these contribution edges are
causally necessary is a counterfactual question;
Section~\ref{sec:limitations} discusses the gap between
contribution and causal effect.


\section{Setup for the token-attribution checks}
\label{app:baselines}

\paragraph{Sentences.}
The $300$ Pile sentences are $8$ to $45$ tokens long, kept when
the model's argmax next token is a confident content word
($p \ge 0.1$, alphabetic). Without this filter, two thirds of the
targets are newline and punctuation defaults, whose likelihood
increases as context is destroyed. Token positions are preserved
during flipping, so the sequence length is constant, and BOS is
excluded from ranking and flipping as a causally inert attention
sink.

\paragraph{Readout.}
We record $\mathrm{softmax}(z)_t$ at each flipping step rather
than the raw logit $z_t$ used by the original protocol. A softmax
model's logits are identified only up to an additive constant:
adding $c$ to every logit leaves the predicted distribution
unchanged. A change in $z_t$ is therefore not by itself a change
in the model's behaviour, since it may consist largely of a
vocabulary-wide shift that the softmax cancels, and it can move in
the opposite direction to the probability the model assigns to the
target. The probability is what the protocol means to measure, and
it is invariant to the shift by construction.

\paragraph{Methods.}
\textbf{IFR} (Information
Flow Routes) is re-implemented from its definition, since the released code is licensed
CC BY-NC; its proximity score is
$\max(\lVert y \rVert_1 - \lVert z - y \rVert_1, 0)$, non-negative by construction and
target-agnostic. \textbf{AttnLRP} and \textbf{CP-LRP} use the authors' LXT library. Each
method runs in its own process, since LXT patches the model globally.

\paragraph{Target direction.}
Each method attributes backward from a single number at the output, a scalar readout of the
target token, and we give every method the same one. The choice matters. LRP relevance is
conservative, so the credit it assigns sums to exactly that number; taking GPT-2's raw target
logit (around $-100$) therefore makes the relevance explain that arbitrary offset and inverts the
heatmap: the relevance sum is negative on $99.5\%$ of examples and AttnLRP scores $6\%$ faithful.
We instead use the vocabulary-mean-centered logit $z_t - \bar{z}$, or a logit difference where a
contrastive target exists (IOI; Appendix~\ref{app:edge_cases}). LXT's own GPT-2 module flags this
issue (``GPT2 has negative
logit values \dots we must also explain the softmax to kick out the negative bias'').


\section{S2 suppression across the Pythia-deduped family}
\label{app:suppression}
\label{app:scaling}

Section~\ref{sec:eval_tokens} defines the four credit quantities
and the suppression index $E$ and reports them on GPT-2 small.
Table~\ref{tab:scaling-s2} reports the same S2 comparison across
five Pythia-deduped scales.

\begin{table}[H]
\centering
\caption{S2 suppression across Pythia-deduped scales ($n=100$,
contrastive targets as in Table~\ref{tab:suppression}).}
\label{tab:scaling-s2}
\small
\begin{tabular}{l ccccc}
\toprule
Model & C$\to$C & S2$\to$S & C$\to$B & S1$\to$S & S1--S2 gap \\
\midrule
160M & $+30.9$ & $+1.5$ & $+14.1$ & $+26.2$ & $+24.7$ \\
410M & $+19.9$ & $-18.1$ & $+1.9$ & $+48.8$ & $+66.9$ \\
1.4B & $+15.5$ & $-39.6$ & $+4.4$ & $+41.3$ & $+80.9$ \\
2.8B & $+32.5$ & $+3.4$ & $+5.1$ & $+35.1$ & $+31.7$ \\
6.9B & $+14.7$ & $-29.3$ & $+4.1$ & $+33.9$ & $+63.2$ \\
\bottomrule
\end{tabular}
\end{table}

The S1--S2 gap is positive at every scale. S2 credit is near zero
or negative at every scale except 2.8B, where the gap narrows and
S2 retains moderate positive credit, suggesting a less structured
IOI circuit at this scale.


\section{Sensitivity to \texorpdfstring{$\beta$}{β}}
\label{app:beta}

The \textsc{SafeDenom} floor $\beta$ (Eq.~\ref{eq:safe_denom})
caps per-step amplification when positive and negative
contributions nearly cancel. The knockout scores of
Section~\ref{sec:eval_knockout} are read off the key-side
decomposition directly (Eqs.~\ref{eq:std},~\ref{eq:mlp}), before
the recursion runs, so they are $\beta$-independent by
construction; the paths and the token attribution pass through
\textsc{SafeDenom} at every step. We sweep $\beta$ against the
perturbation-faithfulness metric of Section~\ref{sec:eval_tokens}
(Table~\ref{tab:beta}), using the same probability readout and
area statistic on $15$ short Pile prompts.

\begin{table}[h]
\centering
\caption{Perturbation faithfulness across $\beta$ ($15$ Pile prompts, probability readout;
$\Delta A$ and \%f as in Table~\ref{tab:faithfulness}). At $\beta = 0$ the floor is off,
per-step credit amplifies without bound, and faithfulness collapses to chance. For every
$\beta \ge 0.2$ it is flat, with $\beta = 0.8$ in the middle of the plateau.}
\label{tab:beta}
\small
\begin{tabular}{r cc}
\toprule
$\beta$ & $\Delta A$ & \%f \\
\midrule
0.0 & $+0.002$ &  53 \\
\midrule
0.2 & $+0.046$ & 100 \\
0.4 & $+0.045$ & 100 \\
0.6 & $+0.060$ & 100 \\
0.8 & $+0.046$ & 100 \\
1.0 & $+0.053$ & 100 \\
1.5 & $+0.045$ & 100 \\
2.0 & $+0.045$ & 100 \\
\bottomrule
\end{tabular}
\end{table}

The floor matters only in that it must be on. With $\beta = 0$, a near-cancelling
denominator lets a single step inflate credit arbitrarily and the ranking degrades to
chance ($\%f = 53$). Once $\beta \ge 0.2$ the amplification is capped and faithfulness is
stable, with no trend across the range and $\beta = 0.8$ (our default throughout)
indistinguishable from its neighbours.


\section{Ablation: output-aligned dispatch}
\label{app:configs}

One choice in the recursion is not forced by the structure. Beyond the first step, credit
is split across a component's entries by projecting each entry's gated value onto the
direction the component actually wrote (Eq.~\ref{eq:stage1} with $d = \phi(S)U$). The
alternative is to split by the gate alone, normalized: the attention weight
$\alpha_{h,q,s}$ for a head, the activation $\phi(\text{pre}_j)$ for an MLP. The gate
already says how strongly each entry was selected, so why also ask where the component
wrote? This appendix answers that. Everything else about the decomposition is held fixed.

\paragraph{The attention sink.}
BOS is an attention sink. It receives large attention weight, but its value contributes
almost nothing to what a head writes into the residual stream. Gate-alone dispatch splits
credit by that attention weight, so it hands position 0 a large share. Projecting onto the
realized output (Eq.~\ref{eq:stage1}) instead weights each entry by what it actually wrote,
so the sink falls away on its own, with no BOS filtering anywhere in the pipeline.

We measure the effect on $50$ prompts each from three sources (Table~\ref{tab:sink}): Pile
sentences, factual-recall prompts, and IOI. Gate-alone inflates the BOS credit share on all
three, but how much that matters depends on the input. On short, low-content prompts (IOI,
factual recall) the sink can outrank the real tokens, taking top-1 on $32\%$ and $18\%$ of
the prompts, while on general text enough later content outweighs it and it wins only $4\%$
of the time. Output-aligned dispatch never lets the sink win, on any source.

\begin{table}[H]
\centering
\caption{The BOS attention sink under the two dispatch rules ($50$ prompts each, GPT-2
small). \emph{Share}: mean percentage of positive token-attribution credit on position 0.
\emph{Top-1}: fraction of prompts where position 0 is the single highest-credited position.}
\label{tab:sink}
\small
\begin{tabular}{l cc cc cc}
\toprule
 & \multicolumn{2}{c}{Pile} & \multicolumn{2}{c}{Factual} & \multicolumn{2}{c}{IOI} \\
\cmidrule(lr){2-3}\cmidrule(lr){4-5}\cmidrule(lr){6-7}
Dispatch & share & top-1 & share & top-1 & share & top-1 \\
\midrule
gate alone (activation-weighted) & 9.2\% & 4\% & 22.3\% & 18\% & 28.9\% & 32\% \\
output-aligned (ours)            & 6.5\% & 0\% & 13.0\% &  0\% & 18.8\% &  0\% \\
\bottomrule
\end{tabular}
\end{table}

\end{document}